\documentclass[lettersize,journal]{IEEEtran}
\usepackage{amsmath,amsfonts}
\usepackage{array}
\usepackage[caption=false,font=normalsize,labelfont=sf,textfont=sf]{subfig}
\usepackage{textcomp}
\usepackage{stfloats}
\usepackage{url}
\usepackage{verbatim}
\usepackage{cite}
\usepackage{graphicx}%
\usepackage{multirow}%
\usepackage{amsmath,amssymb,amsfonts}%
\usepackage{amsthm}%
\usepackage{mathrsfs}%
\usepackage{xcolor}%
\usepackage{textcomp}%
\usepackage{manyfoot}%
\usepackage{booktabs}%
\usepackage{algpseudocode}%
\usepackage{listings}%  
\usepackage{amsmath}
\usepackage{multirow}
\usepackage{subcaption}
\usepackage{float}
\usepackage{booktabs}
\usepackage{multirow}
\usepackage{tabularx}
\usepackage{subfig}
\usepackage{subcaption}
\usepackage{adjustbox} 
\usepackage{algorithm}
\usepackage{algpseudocode}   % 而不是 \usepackage{algorithmic}
\begin{document}

\title{MTSSL: Meta-Thresholding Semi-Supervised Learning}

\author{Shuyang Liu, Ziang Zeng, Ruiqiu Zheng, Jiazheng Wang, Zechen Liu, Wenxi Li, Zhou Yu
        % <-this % stops a space
\thanks{Shuyang Liu, Ziang Zeng, Ruiqiu Zheng, Jiazheng Wang, Zechen Liu, Wenxi Li, Zhou Yu are with School of Statistics, East China Normal University, Shanghai 200062, China  (e-mail: 51204404004@stu.ecnu.edu.cn, 10195000410@stu.ecnu.edu.cn, 52294404002@stu.ecnu.edu.cn, 10225001406@stu.ecnu.edu.cn, 10225000435@stu.ecnu.edu.cn, wxli@sfs.ecnu.edu.cn, zyu@stat.ecnu.edu.cn)}% <-this % stops a space
\thanks{Bangwei Liu and Shaohui Lin are with School of Computer Science, East China Normal University, Shanghai 200062, China (e-mail: 51265901118@stu.ecnu.edu.cn, shaohuilin007@gmail.com)}
\thanks{Corresponding author: Zhou Yu}
\thanks{The datasets utilized in this study are all publicly accessible benchmark datasets. The source code will be made available on GitHub following the acceptance of this paper.}
}
%\thanks{Manuscript received April 19, 2021; revised August 16, 2021.}}


\IEEEpubid{0000--0000/00\$00.00~\copyright~2026 IEEE}

\maketitle

\begin{abstract}
A large body of Semi-supervised Learning~(SSL) algorithms 
encounter the threshold $\tau$ to select pseudo-labels. The value of $\tau$ across different SSL algorithms can vary depending on the learning perspective, yet they may achieve similar performance. 
It motivates us to establish a unified theoretical framework to explain the role of $\tau$ in SSL. 
%In this paper, we propose a statistical explanation: the correct and wrong pseudo-labels affect the unsupervised loss separately, and the value of $\tau$ helps to balance the corresponding error term by affecting their numbers. 
We statistically explained that the unsupervised loss is affected independently by correct and incorrect pseudo-labels, while $\tau$ adjusts their numbers to balance the corresponding error term. 
This inherent trade-off indicates
that SSL can reach the same loss with varying $\tau$, precise optimal values of $\tau$ during training may be unnecessary. 
With this, we treat $\tau$ as an updatable parameter and optimize it via differentiation; the new policy is named \textbf{Meta-Thresholding Semi-Supervised Learning (MTSSL)}. 
%Considering that this unsupervised loss term is common across SSL methods, the MTSSL can be seamlessly deployed in existing SSL frameworks. 
%We smooth the unsupervised loss function of SSL so that $\tau$'s update can be obtained by differentiation and deployed in classical SSL algorithms, and we name the new algorithm \textbf{Meta-Thresholding Semi-Supervised Learning~(MTSSL)}. 
% observe the \textbf{collapse solution} in SSL's adaptive thresholding for the first time: $\tau$ tends to a monotonic increase ultimately and rejects all inputs. 
Extensive experiments demonstrate the superior performance of MTSSL. We observe that the accuracy curves of SSL algorithms can overlap completely even when the values of $\tau$ differ significantly, which supports our theoretical framework and indicates that the selection of $\tau$ can be relaxed in the future design of SSL algorithms.
\end{abstract}


\section{Introduction}
\begin{figure*}[ ]
  \centering
   \includegraphics[width=1\linewidth]{ 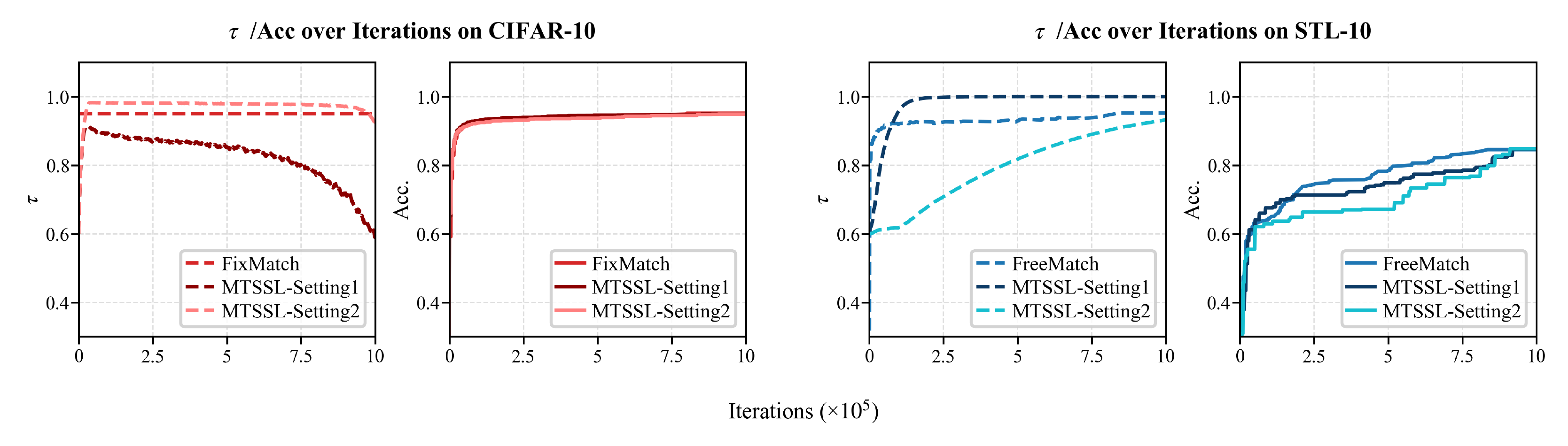}
   \caption{With baseline models FixMatch and FreeMatch, the values of $\tau$ for different algorithms can be significantly different during training. However, the accuracy can be similar or even overlap. The details of setting 1,2 can be found in the Appendix. }
   \label{fig:introduction}
\end{figure*}

\IEEEPARstart{S}{emi}-supervised Learning (SSL) has become increasingly important in deep learning, as it can sufficiently use unlabeled data  \cite{sohn2020fixmatch,wang2022freematch,lee2013pseudo,ijcv2,ijcv3,ijcv4}. 
The two important trends of SSL are consistency regularization and pseudo-labeling: the former encourages the input's high-confidence predictions close to its neighbors  \cite{li2023instant,tarvainen2017mean,rasmus2015semi,berthelot2019remixmatch,ijcv1}, and the latter generates high-confidence pseudo-labels and treats them as true labels  \cite{iscen2019label,zhao2022lassl,zhou2003learning,azizyan2013density}. 
However, the SSL still suffers from \textit{confirmation bias}  \cite{arazo2020ijcnn}: some pseudo-labels or predictions may be wrong and mislead the training. 
Recently, there have been some attempts in SSL to reduce the confirmation bias: 
  \cite{arazo2020ijcnn} pioneered the use of a fixed high-confidence threshold $\tau$ for filtering, considering that high prediction scores are probably correct. 
This filtering strategy has become the dominant paradigm in SSL: the vast majority of SSL algorithms employ a high threshold $\tau$ (usually, $\tau=0.95$ ) to select confidence predictions~\cite{sohn2020fixmatch, zhao2022lassl, zheng2022simmatch, li2021comatch}. 
Although the fixed high threshold $\tau$ can significantly prevent the confirmation bias, excessively high $\tau$ results in fewer pseudo-labels  \cite{guo2022adsh,zhang2021flexmatch,berthelot2022adamatch}, which subsequently hampers SSL's training. 
\iffalse
To select more pseudo-labels, some algorithms aim to sharpen the predictions or pseudo-labels to surpass the threshold $\tau$  \cite{yang2023shrinking, xie2020uda}. 
Some other works maintain the predictions' value but use different class-aware thresholds  \cite{guo2022adsh,guo2022class}. 
\fi
To address this, recent works set a thresholding policy: some algorithms set a low threshold $\tau$ in the early training stage to accept more pseudo-labels, and later increase the value of $\tau$ to select high-confidence predictions  \cite{wang2022freematch,berthelot2022adamatch, xu2022dash}. 
FlexMatch gradually decreasing $\tau$ during some training stages  \cite{zhang2021flexmatch} to help accept more pseudo-labels. 
Counterintuitively, these thresholding policies, with diverse motivations and significantly different values of $\tau$ during training, all demonstrate notable improvements for the SSL. 
In summary, although different values of $\tau$ can yield competitive performance (as shown in Fig.~\ref{fig:introduction}), a more comprehensive theoretical framework is still needed to fully explain its behavior.

In SSL, suppose we have $n$ unlabeled data and every randomly selected input $x_i,~i\in\{1,...,n\}$ has a correct prior label $y_i$. Denote $a(\cdot),A(\cdot)$ as weak and strong augmentations, $p_\theta(x)$ as the prediction with parameter $\theta$, $p'_\theta(x)$ as the the maximize element of $p_\theta(x)$, then most SSL algorithms  \cite{sohn2020fixmatch,zheng2022simmatch,zhao2022lassl,chen2023softmatch,nguyen2023boosting} incorporate the empirical unsupervised loss $\widehat L_u(\theta,\tau)$:       
\begin{equation}
\label{Lu}
     \widehat L_u(\theta, \tau)=\frac{1}{n}\sum_{i=1}^nI\big(p'_{\theta}(x_i)\geq \tau\big)\cdot H\big[p_\theta(A(x_i)),p_\theta(a(x_i))\big],
\end{equation}  
where we define 
\begin{equation*}
H\big[p_\theta(A(x_i)),p_\theta(a(x_i))\big] = -p_\theta(A(x_i))\cdot\log\big(p_\theta(a(x_i))\big).
\end{equation*}
By selecting as many as possible correct predictions with a threshold $\tau$, Eq.(\ref{Lu}) aims to encourage the model's consistency and extract better representations\cite{sohn2020fixmatch,wang2022freematch,berthelot2019remixmatch,zhao2022lassl,sosea2023marginmatch}. 
Denote $y^\theta_i$ as the artificial one-hot label derived from $p'_\theta(x_i)$, SSL algorithms use the Eq.(\ref{Lu}) to replace the 
\begin{equation*}
     \widehat L_u^*(\theta)=\frac{1}{n}\sum_{i=1}^nI\big(y^\theta_i=y_i\big)\cdot H\big[p_\theta(A(x_i)),p_\theta(a(x_i))\big],
\end{equation*}  
which can fully use the unlabeled data with prior labels $y_i$. 

Suppose the random variable of input data is $(X,Y)$, where $Y$ is the prior one-hot label, and $Y^\theta$ is the artificial one-hot label w.r.t. the prediction $p_\theta(X)$. 
Then, the population version of $\widehat L_u^*(\theta)$ is: 
\begin{equation}
\label{2}
     L_u^*(\theta)=\mathbb{E}_{(X,Y)}\big[I(Y^\theta = Y)\cdot H\big[p_\theta(A(X)),p_\theta(a(X))\big]\big],
 \end{equation}
and SSL algorithms try to estimate $L_u^*(\theta)$ by $\widehat L_u(\theta, \tau)$ in Eq.(\ref{Lu}). 
Then, SSL algorithms with a lower estimation error
\begin{equation}
\label{3}
    \big|L_u^*(\theta)-\widehat L_u(\theta, \tau)\big|
\end{equation}
yields a better estimation. 
However, the predictions satisfy $p'_\theta(x)\geq \tau$ may contain both correct and wrong pseudo-labels \cite{arazo2020ijcnn}, misleading the value of Eq.(\ref{Lu}) and thus the estimation error. Denote $\widehat L_u^{n_C}(\theta,\tau)$ represents the part with correct pseudo-labels, and $\widehat L_u^{n_W}(\theta,\tau)$ the part with wrong pseudo-labels, we have: 
\begin{equation*}
    \widehat L_u(\theta,\tau) =\widehat L_u^{n_C}(\theta,\tau) + \widehat L_u^{n_W}(\theta,\tau).
\end{equation*}
Then the estimation error $|L_u^*(\theta) - \widehat L_u(\theta,\tau)|$ is influenced by two error terms: $ L_u^*(\theta) - \widehat L_u^{n_C}(\theta,\tau)$ and $\widehat L_u^{n_W}(\theta,\tau)$, the value of which are determined by the numbers of correct and incorrect pseudo-labels, respectively. 
Intuitively, we hope to select more correct pseudo-labels and fewer wrong pseudo-labels to get a lower estimation error. 
We explain theoretically that the estimation error exhibits an inherent trade-off with the threshold $\tau$, as $\tau$ directly influences the number of correct and wrong pseudo-labels. 
This theoretical framework unifies several previous SSL thresholding policies with quite different values of $\tau$  \cite{wang2022freematch, zhang2021flexmatch, berthelot2022adamatch}, pointing out that different values of $\tau$ can achieve favorable results by balancing the two error terms $L^*_u(\theta) - \widehat L_u^{n_C}(\theta,\tau)$ and $\widehat L^{n_W}(\theta,\tau)$.

%The previous SSL adaptive thresholding policies lack full consideration of $\tau$'s impact, and our theoretical framework reveals that different values of $\tau$ may bring the same $\widehat L_u(\theta,\tau)$. 
The previous SSL adaptive thresholding policies provide different $\tau$ values for various perspectives, and our theoretical framework reveals that different values of $\tau$ may bring the same $\widehat L_u(\theta,\tau)$. 
It motivates us to give up providing precise $\tau$ values during training but treat $\tau$ as a parameter and optimize it directly by differentiation, and name our thresholding policy as Meta-Thresholding Semi-Supervised Learning (MTSSL). 
To elegantly deploy MTSSL to a large number of traditional SSL algorithms, it keeps the common unsupervised loss $\widehat L_u(\theta,\tau)$ and replaces the non-differentiable indicator function $I(p'_{\theta}(x_i) \geq \tau)$ with the smooth sigmoid function\cite{byod,nestrov}. 
However, with the sigmoid function, the gradient of $\tau$ will be positive and result in a degenerate solution $\tau > 1$, yet rejects all predictions and stops the update.
We name it the \textit{collapse solution}, and a regularizer is added to control the value of $\tau$. 
Extensive experiments show that MTSSL achieves better performance than the existing best SSL thresholding policy, FreeMatch \cite {wang2022freematch}. 
We also deploy the MTSSL policy to FixMatch, the classical SSL baseline model with a fixed threshold, and observe a significant improvement. 
%Specifically, MTSSL exceeds the baseline model with a \textbf{2.64\%} improvement over the STL-10 dataset with 40 labeled examples. 
We observe that the accuracy curves of SSL algorithms may almost overlap with different values of $\tau$ during training, which inspires us that SSL algorithms may not be so sensitive to the threshold $\tau$. 

Our contributions can be summarized as:

• A statistical framework is proposed to explain the effect of $\tau$ in SSL for the first time. We reveal that the estimation error $|L_u^*(\theta) - \widehat L_u(\theta,\tau)|$ is affected by the value of $\tau$ inherently, for $\tau$ controls the number of correct and wrong pseudo-labels;

• We design a plug-and-play thresholding algorithm, MTSSL. MTSSL can be easily deployed to various SSL algorithms and automatically update the value of $\tau$; 

• We conduct extensive experiments on MTSSL. On the STL-10 and CIFAR-100 datasets, it surpasses the baseline with improvements of $\textbf{2.64}\%$ and $\textbf{1.51}\%$. 
Our experiments validate the theoretical analysis, which reveals that SSL exhibits limited sensitivity to $\tau$ and inspires us to relax the selection of $\tau$ in the future.

\section{Related Works}
\subsection{Semi-supervised Learning(SSL)}

There are two main trends in SSL: Pseudo-labeling and consistency regularization. 
Pseudo-labeling focuses on generating high-quality pseudo-labels. Recent frameworks construct more complex pseudo-labels to improve their quality: Comatch  \cite{li2021comatch} introduces graphic information into contrastive learning and smooths pseudo-labels with a memory bank; Simmatch  \cite{zheng2022simmatch} optimizes pseudo-labels by aligning their marginal distributions. 
Label propagation  \cite{zhou2003learning,iscen2019label, zhao2022lassl, douze2018low} iteratively propagates the pseudo-label with neighbors' features. 
Consistency regularization expects a good encoder to output near predictions for near inputs. 
$\pi$-model firstly expects the predictions between the images and their augmentations to be similar. 
Mixup or adversarial training can replace the augmentation and produce similar images for SSL algorithms \cite {berthelot2019remixmatch,miyato2018virtual}. 
LASSL  \cite{zhao2022lassl} introduces contrastive learning and limits the comparisons within the same category. 

However, SSL algorithms still suffer from confirmation bias: they inevitably contain wrong pseudo-labels or predictions  \cite{arazo2020ijcnn, sosea2023marginmatch}.  
There are some denoising techniques in SSL: 
Marginmatch  \cite{sosea2023marginmatch} dynamically selects pseudo-labels by considering their behavior during the whole training; 
Regmixmatch  \cite{han2025regmixmatch} mixup low-confidence samples to generate extra labels, with which to potentially counteract the impact of over-confident predictions; Adamatch  \cite{berthelot2022adamatch} only chooses a percentile of generated pseudo-labels. 
However, these algorithms can't proactively generate high-quality pseudo-labels. 
  \cite{chen2023boosting} aims to enhance the quality of pseudo-labels by considering the early exclusion of erroneously classified categories.
Dividemix independently trains two SSL networks to select better pseudo-labels  \cite{li2020dividemix}. 
Considering these algorithms incur additional computational costs, the main tendency in SSL to reduce the confirmation bias is still simply setting a filtering threshold $\tau$  \cite{arazo2020ijcnn, sohn2020fixmatch, berthelot2019mixmatch, zhao2022lassl}. 
%In SSL, the main tendency to reduce the \textit{confirmation bias} is to set a filtering threshold $\tau$  \cite{arazo2020ijcnn, sohn2020fixmatch, berthelot2019mixmatch, zhao2022lassl}, but this strategy is criticized for giving up too much unlabelled data  \cite{yang2023shrinking, han2025regmixmatch}. 
%In a word, we still need a simple yet elegant design with a theoretical explanation. 

\subsection{Adaptive Thresholding In SSL}
Although many SSL algorithms reduce the confirmation bias by setting a fixed high threshold  \cite{arazo2020ijcnn,sohn2020fixmatch,zheng2022simmatch,zhao2022lassl}, this strategy is sometimes criticized for ignoring too many low-confidence predictions  \cite{yang2023shrinking, han2025regmixmatch}. 
Freematch  \cite{wang2022freematch} gradually updates the $\tau$ by historical confidence to balance the need for pseudo-labels in different training stages and categories. Regmixmatch  \cite{han2025regmixmatch} inherits this tuning policy when generating extra pseudo-labels. 
Flexmatch  \cite{zhang2021flexmatch} gradually reduces $\tau$ to accept more pseudo-labels. 
%Adamatch  \cite{berthelot2022adamatch} constructs a relative threshold with distribution alignment and addresses the distribution shift. 
Adsh  \cite{guo2022adsh} setting different adaptive thresholds for different classes and thus encouraging a class-balanced prediction. 
Instant  \cite{li2023instant} constructs a noise transition matrix to correct the labels, but the computational cost is too high. 
Some SSL algorithms try to incorporate \textit{meta-learning}  \cite{duchi2011adagard,2022mlr}, by which we can predict the threshold $\tau$ or pseudo-labels during the training process: 
  \cite{xiao2021aistat} use meta-learning to improve pseudo-labels. 
%Several SSL algorithms try to incorporate \textit{meta-learning}  \cite{duchi2011adagard,2022mlr}, by which we can predict hyperparameters during the training process:   \cite{xiao2021aistat} uses meta-learning to improve pseudo-labels, and   \cite{zhang2023mlh} uses meta-learning to deal with time-series data. 
Dash  \cite{xu2022dash} predicts $\tau$ in a bi-level optimization framework, and calculates the precise value of $\tau$ in different stages, but its performance is unsatisfactory. 
However, although these algorithms differ in motivations and utilize distinct $\tau$ values during training, they show notable improvements accordingly. 
The impact of the threshold $\tau$ for SSL is still ambiguous; we urgently require a unified theoretical framework to clarify the role and nature of $\tau$ in SSL.

\section{Method}

\subsection{The Inherent Trade-off of $\tau$}
Even the high-confidence predictions satisfies $p'_\theta(x_i)\geq \tau$ inevitably contains some wrong pseudo-labels~\cite{arazo2020ijcnn}, thus affects the estimation error $|L_u^*(\theta)-\widehat L_u(\theta, \tau)|$ by misleading the $\widehat L_u(\theta, \tau)$. 
%We analyze the impact of correct and wrong pseudo-labels, respectively. 
Suppose we have $n_C$ correct pseudo-labels and $n_W$ wrong pseudo-labels, denote $y^\theta_i$ as the artificial one-hot label w.r.t. the predicted $p'_\theta(x_i)$, define $n_C,n_W$ as the number of correct and wrong pseudo-labels:
\begin{equation*}
      n_C=\operatorname{Card}\big(\{x_i: p'_\theta(x_i)\geq \tau,y^\theta_i = y_i,i=1,...,n\}\big),
\end{equation*}
\begin{equation*}
      n_W = \operatorname{Card}\big(\{x_i: p'_\theta(x_i)\geq \tau,y^\theta_i \neq y_i,i=1,...,n\}\big), 
\end{equation*} 
where $\operatorname{Card}(\cdot)$ represents the number of elements of a set. 
To simplify the notation, denote the event $A, B$ as
\begin{equation*}
    A=\big\{x_i\big|p'_\theta(x_i)\geq \tau ,y^\theta_i = y_i, i=1,...n\big\},
\end{equation*}
\begin{equation*}
    B=\big\{x_i\big|p'_\theta(x_i)\geq \tau,y^\theta_i \neq y_i, i=1,...n\big\}.
\end{equation*}
Then we can define: 
\begin{equation*}
    \widehat L_u^{n_C}(\theta,\tau) = \frac{1}{n}\sum_{i=1}^{n}I\big(x_i \in A\big)\cdot H\big[p(A(x_i)),p(a(x_i))\big],
\end{equation*}
\begin{equation*}
    \widehat L_u^{n_W}(\theta,\tau) = \frac{1}{n}\sum_{i=1}^{n}I\big(x_i\in B\big)\cdot H\big[p(A(x_i)),p(a(x_i))\big],
\end{equation*}
with which we can decompose the $\widehat L_u(\theta,\tau)$ in Eq.(\ref{Lu}) as 
\begin{equation*}
    \widehat L_u(\theta,\tau)= \widehat L_u^{n_C}(\theta,\tau) + \widehat L_u^{n_W}(\theta,\tau).
\end{equation*}

We hope the empirical unsupervised loss $\widehat L_u(\theta,\tau)$ is close to the population loss $L_u^*(\theta)$ as much as possible, and the estimation error can be decomposed as: 
\begin{equation}
    \big|L_u^*(\theta) - \widehat L_u(\theta,\tau)\big| = \big|\big(L_u^*(\theta)  - \widehat L_u^{n_C}(\theta,\tau)\big) - \widehat L_u^{n_W}(\theta,\tau)\big|.
    \label{Eq.3}
\end{equation}

Intuitively, we expect a larger $n_C$ to decrease $|L_u^*(\theta)  - \widehat L_u^{n_C}(\theta,\tau)|$ and a smaller $n_W$ to decrease the term $\widehat L_u^{n_W}(\theta,\tau)$, for more correct predictions brings a closer estimation to $L^*_u(\theta)$, while less wrong predictions results in a smaller $\widehat L_u^{n_W}(\theta,\tau)$. 
We statistically illustrate the intuition following a classical setting  \cite{wang2022freematch, yang2020rethinking}. Suppose the input random variable $(X,Y)$ has the following conditional normal distribution:
\begin{equation*}
    X \mid_{Y = -1} \sim N(\mu_1, \sigma_1^2),~ X \mid_{Y = +1} \sim N(\mu_2, \sigma_2^2), ~~\mu_2 > \mu_1.
\end{equation*}
And we adopt the classical logistic model as the encoder with parameter $\theta \in \mathbb{R}^+$: 
\begin{equation*}
    p_\theta(x) = 1/\big[1+\exp\big(-\theta\cdot(x-\frac{\mu_1+\mu_2}{2})\big)\big].
\end{equation*}
A sample $x$ is assigned a pseudo-label $+1$ if $p_\theta(x) > \tau$ and $-1$ if $p_\theta(x) < 1 - \tau$. The pseudo label is 0 if $ p_\theta(x) \in [1-\tau, \tau], \tau \in (1/2,1]$.  
Under these assumptions, we have: 

\par
~
\par
\textbf{Proposition 1}
\textit{For Logistic Model with parameter} $\theta$ \textit{and threshold} $\tau$, introduce $\tau' = \frac{1}{\theta}\log(\frac{\tau}{1-\tau})$ \textit{to simplify the notation.}
\textit{When given $n$ unlabeled data, the number of wrongly-assigned pseudo-labels is: }
\begin{equation}
\label{Eq5.13}
    n_W=  n\cdot\bigg(1- \Phi\bigg(\frac{\frac{\mu_2-\mu_1}{2}+\tau'}{\sigma_1}\bigg)\bigg) +  n\cdot \Phi\bigg({\frac{\frac{\mu_1-\mu_2}{2} - \tau'}{\sigma_2}\bigg)},
\end{equation}
\begin{equation}
\label{Eq5.14}
n_C = n\cdot\bigg(1-\Phi\bigg(\frac{\frac{\mu_1-\mu_2}{2}+\tau'}{\sigma_2}\bigg)\bigg) +  n\cdot \bigg(\Phi\bigg(\frac{\frac{\mu_2-\mu_1}{2}-\tau'}{\sigma_1}\bigg)\bigg).
\end{equation}
\textit{The value of $n_W, n_C$ monotonically decreases with the value of $\tau$. }
\par
~
\par

The Eq.(\ref{Eq5.13}) and Eq.(\ref{Eq5.14}) indicate that the value of $\tau$ can significantly affect the number of both correct and wrong pseudo-labels, and both $n_C, n_W$ monotonically decrease with $\tau$. 
Specifically, when $\tau \rightarrow 1$, we have $n_C, n_W \rightarrow 0$, and thus $P_C, P_W \rightarrow 0$, where $P_C=\frac{n_C}{n}$, $P_W=\frac{n_W}{n}$. 
Next, we analyze the impact of the number of correct and incorrect pseudo-labels in \textbf{Theorem 2} and \textbf{Theorem 3}. 

\par
~
\par
\textbf{Theorem 2}
\textit{Under classical assumptions\cite {sub-gaussian3, sub-gaussian2}, $\widehat L_u^{n_C}(\theta,\tau, X)$ is a sub-Gaussian random variable with parameter $\gamma, C_1$. Denote $P_C = \frac{n_C}{n}$, then for $\forall t \in \mathbb{R}^+$, we have: }
\begin{small}
\begin{equation}
\label{Eq5.9}
    P\left(  \left| \widehat{L}_u^{n_C}(\theta,\tau) -P_C\cdot L^*_u(\theta)  \right|  > t \right) \leq \exp\Bigl(-\frac{n^2 t^2}{2\gamma n_C}\Bigr).
\end{equation}
\end{small}
Here $P_C$ monotonically decreases with the value of $\tau$. When $\tau \rightarrow 1$, $P_C\rightarrow 0$, we have $|\widehat{L}_u^{n_C}(\theta,\tau)-L^*_u(\theta)| \rightarrow L^*_u(\theta)$.
\par
~
\par

With Eq.(\ref{Eq5.9}), when we have $n\rightarrow +\infty$, for any $t\in \mathbb{R}^+$, it's obvious that $-\frac{n^2t^2}{2\gamma n_C} \rightarrow -\infty$. Then the Eq.(\ref{Eq5.9}) indicates that $|\widehat{L}_u^{n_C}(\theta,\tau)- P_C\cdot L^*_u(\theta) >t| \rightarrow 0$ holds for any $t\in \mathbb{R}^+$. 
Considering that $P_C$ monotonically decrease with the value of $\tau$, when $\tau\rightarrow 0$, $P_C$ tends to increase, and $\widehat{L}_u^{n_C}(\theta,\tau)$ yielding a better estimation of $L^*_u(\theta)$. 
However, it's too early to state that SSL prefers a smaller threshold $\tau$, for we still need to analyze the behavior of wrong pseudo-labels. 

\par
~
\par
%\cite{gaussian_augmentation}
\textbf{Theorem 3}
\textit{Following the classical setting  }, \textit{we suppose the augmentation $a(x),A(x)$ for any input $x$ by adding the Gaussian-random noise to $x$  , that is: }
\begin{equation*}
    a(x)=x+\delta_1,~~~A(x)=x+\delta_2,~~~\delta_1,\delta_2\sim N(0,\sigma_3).
\end{equation*}
\textit{Then when $n \rightarrow +\infty$,  with probability 1, we have: $\widehat L_u^{n_W}(\theta,\tau) \propto O(\frac{n_W}{n}).$ }
\par
~
\par

We have deduced that $n_W$ decreases with the value of $\tau$.
When $\tau \rightarrow 0$, with $\widehat{L}_u^{n_C}(\theta,\tau)$ tends to approach $L^*_u(\theta)$, $\widehat{L}_u^{n_W}(\theta,\tau)$ results in remarkable confirmation-bias and misleads the SSL's training. 
When $\tau \rightarrow 1$, $\widehat{L}_u^{n_C}(\theta,\tau), \widehat{L}_u^{n_W}(\theta,\tau) \rightarrow 0$ for the same time. 
Although there are few wrong pseudo-labels, we can not provide a good estimation of $L^*_u(\theta)$ due to the lack of correct pseudo-labels. 
In a word, SSL's estimation error in Eq.(\ref{3}) has an inherent trade-off with the threshold $\tau$, either lower or higher $\tau$ may reach the same value of $|L_u^*(\theta)-\widehat L_u(\theta,\tau)|$ by balancing the value of terms $\widehat L_u^{n_W}(\theta,\tau)$, $L_u(\theta)  - \widehat L_u^{n_C}(\theta,\tau)$ respectively. 
\textbf{Theorem 2} and \textbf{Theorem 3} can theoretically explain the counter intuitively phenomenon shown in Fig.(\ref{fig:introduction}).

\subsection{Meta-learning Based Thresholding Policy}
\textbf{Theorem 2, 3} reveals that different values of $\tau$ may lead to the same estimation error $|L_u^*(\theta) - \widehat{L}_u(\theta, \tau)|$ in Eq.(\ref{3}); it isn't necessarily to find an exact optimal threshold $\tau$ during training for SSL algorithms. 
%It motivates us to give up the case-by-case predefined thresholding policies and replace them with a much simpler and easier-to-implement policy for any SSL algorithms. 
It motivates us that different values of $\tau$ may bring the same unsupervised loss and thus the same training effect.
It's possible to replace the case-by-case predefined thresholding policies with a more intuitive and easier-to-implement policy. 
%We name our new policy Meta-Thresholding Semi-Supervised Learning (MTSSL). 
The simplest yet elegant approach is to view the threshold $\tau$ as a parameter and optimize it by back propagation, but the classic SSL algorithms have an unsupervised loss term $\widehat L_u(\theta,\tau)$ as in Eq.(\ref{Lu}) that contains a non-differentiable indicator function involving $\tau$ \cite{sohn2020fixmatch, wang2022freematch, zhang2021flexmatch, zheng2022simmatch, zhao2022lassl}. 
For both fair comparison and the broad applicability of MTSSL, we inherit the classic unsupervised loss term $\widehat L_u(\theta,\tau)$ and smooth the indicator function $I(p'_\theta(x)\geq\tau)$ in $\widehat L_u(\theta,\tau)$ following the classical technique  \cite{byod, nestrov} by substituting the indicator function with a sigmoid function and get the smoothed unsupervised loss $\widehat L_u'(\theta,\tau)$:
\begin{equation}
\label{sigmoid}
    \widehat L_u'(\theta,\tau)=\sum_{i=1}^n  \frac{1}{1+e^{-\beta\cdot(p'_\theta(x_i)-\tau)}}\cdot H\big[p_\theta(a(x_i)),p_\theta(A(x_i)\big].
\end{equation}

We observe that during the optimization of Eq.(\ref{sigmoid}), the threshold $\tau$ tends to increase sharply until $\tau > 1$. 
The high $\tau$ reduces the $\widehat L'_u(\theta,\tau)$ meaningless to 0 by decrease $n_C=n_W=0$, for $p'_\theta(x)\leq1$. 
We name the phenomenon the \textit{collapse solution} and provide a simple explanation: 

\par
~
\par
\textbf{Proposition 4}\label{prop4}
\textit{For $\widehat L_u'(\theta, \tau)$ by replacing the indicator function with a sigmoid function in Eq.(\ref{sigmoid}), with any $\beta \in \mathbb{R}^+$: }
\begin{equation*}
    \widehat L_u'(\theta,\tau) = \sum_{i=1}^n  \frac{1}{1+e^{-\beta\cdot((p'_\theta(x_i)-\tau))}}\cdot H\big[p_\theta(A(x_i)),p_\theta(a(x_i))\big],
\end{equation*}
\textit{we always have:} $\frac{\partial \widehat L'_u(\theta,\tau)}{\partial \tau} <0$.
\par
~
\par

\begin{algorithm}
\caption{The algorithm of MTSSL}
\label{pseudo_code}
\begin{algorithmic}[1]
\State \textbf{Input:} $n$ unlabeled data, initial model parameter $\theta_0$,
      initial threshold $\tau_0$, hyperparameter $K$, regularizer $g(\cdot)$,
      regularization weight $\lambda$, $\eta_1$ (learning rate for $\theta_t$),
      $\eta_2$ (learning rate for $\tau_t$).

\State \textbf{Output:} Current model parameter $\theta_t$, current smoothed threshold $h(\tau_t)$.

\State Update model parameter with SGD:
\[
    \theta_{t+1} = \theta_t - \eta_1 \cdot \frac{\partial \big[L_s + \widehat{L}_u'(\theta_t, h(\tau_t))\big]}{\partial \theta_t}
\]

\If{$t \bmod K = 0$}
    \State Update the current threshold $\tau_t$:
    \[
        \tau_{t+1} = \tau_t - \eta_2 \cdot \frac{\partial \left( \widehat{L}_u'(\theta_t, h(\tau_t)) \right)}{\partial \tau_t}
    \]
    \State Output the smoothed threshold and model parameter: $h(\tau_{t+1})$, $\theta_t$.
\EndIf
\end{algorithmic}
\end{algorithm}

\textbf{Proposition 4} indicates that with the smoothed loss term $\widehat L'_u(\theta,\tau)$ in Eq.(\ref{sigmoid}), $\tau$ tends to increase monotonically by differentiation. 
Although the collapsed solution may be avoided by designing a new unsupervised loss, we choose to smooth the indicator function involving the threshold $\tau$ for a fair comparison, so that our thresholding policy can be easily embedded into various classic SSL algorithms that incorporate the classical unsupervised loss $\widehat{L}_u(\theta, \tau)$ in Eq.(\ref{Lu})  \cite{zhang2021flexmatch,wang2022freematch,sohn2020fixmatch, zhao2022lassl, zheng2022simmatch}. 
To control the collapse solution, we introduce 
\begin{equation*}  
h(\tau) = \frac{1}{1+e^{-\tau}}
\end{equation*} 
to replace the original threshold $\tau$, which maps the threshold $\tau$ within the range $(0, 1)$ smoothly. 
We also add a regularizer $g(\cdot)$ to punish excessively near values of $h(\tau)$ to 1. 
Then we construct the new unsupervised loss: 
\begin{equation}
\label{regularizer}
    \widehat L_u'(\theta,\tau) = \sum_{i=1}^n  \frac{H\big[p_\theta(A(x_i)),p_\theta(a(x_i))\big]}{1+e^{-\beta\cdot(p'_\theta(x_i)-h(\tau))}}  + \lambda\cdot g(h(\tau)),
\end{equation}
where $\lambda\in \mathbb{R}^+$ is a weight for the regularizer. 
In this paper, we provide two choices of $g(\cdot)$:
\begin{equation*}  
\left\{  
\begin{aligned}  
&g(h(\tau)) = \|h(\tau)\|_2^2, \\  
&g(h(\tau)) = \sqrt{\frac{1}{1 - h(\tau)}}.  
\end{aligned}  
\right.  
\end{equation*}
The former choice employs a common $l_2$-regularizer to delay the increase of $h(\tau)$, which may benefit the future statistical study. 
The second choice imposes a progressive penalty on $h(\tau)$: when $h(\tau)$ gets closer to 1, the penalty becomes more intense. 
We name our adaptive thresholding algorithm Meta-Thresholding Semi-Supervised Learning (MTSSL). 
The pseudo-code is illustrated in the Algorithm.\ref {pseudo_code}.

\subsection{MTSSL + FreeMatch: Loss Function}
We aim to demonstrate that the simple, plug-and-play MTSSL can replace classical SSL thresholding policies and reach a better result, and significantly improve the fixed-threshold classical SSL algorithms. 
To this end, we choose the SOTA SSL thresholding policy, FreeMatch, and replace its thresholding mechanism with our proposed MTSSL. 
For a fair comparison, we keep FreeMatch's structure to improve imbalanced pseudo-labeling, and only replace the global thresholding policy of FreeMatch. 
Denote the $h(\tau_t)$ as the smoothed threshold in the $t-th$ iteration optimized by MTSSL. 
Then, following the design of FreeMatch, suppose input $x$ has $C$ categories, define $\tau'_t \in \mathbb{R}^C$ as the class-wise threshold, we can calculate the local threshold for the $c$-th class, $\tau_{t,c}'$, w.r.t. the current global threshold $h(\tau_t)$: 
\begin{equation*}
    \tau_{t,c}' = h(\tau_t)\cdot\frac{\widetilde p_{t,c}}{\max(\widetilde p_{t,c}:c\in[C])},
\end{equation*}
%where we have $\tau_{0,c}' = 1/C$,$C$ is the number of class and define: 
where we define:
\begin{equation*}
    \widetilde p_{t,c} = \eta\cdot \widetilde p_{t-1,c} +  (1-\eta)\cdot\frac{\sum_{i=1}^{n}p_{t,c}(x_i)}{n},
\end{equation*}
here we denote $p_{t,c}(x_i)$ as the $c$-th element of the prediction $p_{\theta_t}(x_i)$. 
Then we can define the unsupervised loss:
\begin{equation*}
    \widehat L_u =     \frac{1}{n}\cdot\sum_{i=1}^n  H\big[p_\theta(A(x_i)),p_\theta(a(x_i))\big]\big|_{p_\theta'(x_i)>\tau_{t,c_{i,\max}}'},
\end{equation*}
here we note $c_{i,\max}$ as the category of the largest prediction $p'_\theta(x_i)$. 
Then we further normalize the expectation of probability by the histogram
distribution of pseudo labels selected by $h(\tau_t)$. Define:
\begin{equation*}
    \widetilde h_t= \eta\cdot \widetilde h_{t-1} + (1-\eta)\cdot \operatorname{Hist}_n|_{p'_\theta(a(x_i))\geq \tau_{t,c_{i,\max}}'},
\end{equation*}
where $\operatorname{Hist_n}(\cdot)$ means the history gram of $n$ samples' prediction. 
Define $\operatorname{SumNorm}=(\cdot)/\sum(\cdot)$, and 
\begin{equation*}
    \overline p_t = \sum_{i=1}^n p_\theta(a(x_i))|_{p'_\theta(x_i)\geq \tau_{t,c_{i,\max}}'},
\end{equation*}
\begin{equation*}
    \overline h = \operatorname{Hist}_n|_{ p'_\theta(x_i)\geq \tau_{t,c_{i,\max}}'}.
\end{equation*}
Then we define the consistency loss: 
\begin{equation*}
    \widehat L_f = H\bigg(\operatorname{SumNorm}\bigg(\frac{h(\tau_t)}{\widetilde h_t}\bigg),\operatorname{SumNorm}\bigg(\frac{\overline p_t}{\overline h_t}\bigg)\bigg),
\end{equation*}
Then we finish the insertion of MTSSL into Freematch and get the loss function: 
\begin{equation*}
    L = \widehat L_s + \widehat L_u + \widehat L_f.
\end{equation*}
We show the superior performance of this algorithm in experiments. In fact, MTSSL+FreeMatch yields the best adaptive thresholding policy for semi-supervised learning.

\subsection{MTSSL + FixMatch: Loss Function}
We also demonstrate that the simple, plug-and-play MTSSL policy can achieve significant improvements over fixed-threshold SSL algorithms. 
We adopt FixMatch  \cite{sohn2020fixmatch}, one of the most classical fixed-threshold SSL frameworks, and replace its fixed $\tau=0.95$ with our proposed MTSSL. 
Denote $h(\tau_t)$ as the threshold in the $t$-th iteration optimized by MTSSL; then we first replace the indicator function in $\widehat L_u(\theta,\tau)$ with $h(\tau_t)$ : 
\begin{equation*}
    \widehat L_u'=\sum_{i=1}^n  \frac{H\big[p_\theta(a(x_i)),p_\theta(A(x_i)\big]}{1+e^{-\beta\cdot((p'_\theta(x_i)-h(\tau_t)) }} + \lambda \cdot g(h(\tau_t)).
\end{equation*}
Then, following the loss function of FixMatch, we make full use of the $N$ labeled data $(x_i,y_i)$ with the following supervised loss: 
\begin{equation*}
    \widehat L_s'=\sum_{i=1}^N  H\big[p_\theta(x_i),y_i\big].
\end{equation*}
Then we finish the insertion of MTSSL into FixMatch and get the loss function in the t-th iteration: 
\begin{equation*}
    L_t = \widehat L_s' + \widehat L_u'.
\end{equation*}
The application of MTSSL to FixMatch (Tab.~\ref{tabel1}) brings clear improvements and attains competitive results with recent SOTA in 6 out of 8 settings, which underscores its potential as a plug-and-play strategy that can be readily applied to any classical SSL algorithm that involves an unsupervised loss $\widehat L_u(\theta,\tau)$.

\begin{table*}
\caption{Performance comparison on CIFAR10, CIFAR100, SVHN, and STL10. 
We report mean accuracy (\%) and standard deviation over 5 runs. 
\textbf{Bold} and \underline{underlined} numbers indicate the best and second-best results, respectively.}
\label{tabel1}
\centering
\footnotesize   % 正常字号，缩放后会等比例缩小，但不会像4.5pt那么夸张
\setlength{\tabcolsep}{4pt}  % 列间距稍大，避免缩放后数字贴在一起
\renewcommand{\arraystretch}{1.5}

\begin{adjustbox}{width=\textwidth}   % 整体缩放到页面宽度
\begin{tabular}{@{}l cccccccc@{}}
\toprule
\hline
\multirow{2}{*}{Method} & \multicolumn{2}{c}{CIFAR-10} & \multicolumn{2}{c}{CIFAR-100} & \multicolumn{2}{c}{SVHN} & \multicolumn{2}{c}{STL-10} \\
\cmidrule(r){2-3} \cmidrule(r){4-5} \cmidrule(r){6-7} \cmidrule(r){8-9}
& 40 labels & 250 labels & 400 labels & 2500 labels & 40 labels & 250 labels & 40 labels & 1000 labels \\
\midrule
Pseudo-label (2013)      & 25.39±0.26 & 53.51±2.20 & 12.55±0.85 & 42.26±0.28 & 25.39±5.60 & 79.79±1.09 & 25.31±0.99 & 67.36±0.71 \\
Mean-Teacher (2017)      & 29.91±1.60 & 62.54±3.30 & 18.89±1.44 & 54.83±1.06 & 33.81±3.98 & 96.43±0.11 & 29.28±1.45 & 66.10±1.37 \\
MixMatch (2019)          & 63.81±6.48 & 86.37±0.59 & 32.41±0.66 & 60.24±0.48 & 69.40±3.39 & 96.02±0.23 & 45.07±0.96 & 88.30±0.68 \\
ReMixMatch (2019)        & 90.12±1.03 & 93.70±0.05 & 57.25±1.05 & 73.97±0.35 & 79.96±3.13 & 97.08±0.48 & 67.88±6.24 & 93.26±0.14 \\
UDA (2020)               & 89.33±3.75 & 94.84±0.06 & 53.61±1.59 & 72.27±0.21 & 94.88±4.27 & 98.01±0.02 & 62.58±3.44 & 93.36±0.17 \\
FixMatch (2020)          & 92.53±0.28 & 95.14±0.05 & 53.58±0.82 & 71.97±0.16 & 96.19±1.18 & 98.03±0.01 & 64.03±4.14 & 93.75±0.33 \\
Dash (2021)              & 91.07±3.11 & 94.84±0.23 & 55.18±0.96 & 72.85±0.22 & 97.81±0.18 & 97.96±0.02 & 65.48±4.30 & 93.61±0.56 \\
FlexMatch (2021)         & 95.03±0.06 & 95.02±0.09 & 60.06±1.02 & 73.51±0.20 & 91.81±3.20 & 93.41±2.29 & 70.85±4.16 & 94.23±0.18 \\
SimMatch (2022)          & 94.62±0.01 & 94.64±0.08 & 60.68±0.72 & 73.79±0.37 & 92.40±2.11 & 97.95±0.05 & 83.02±4.24 & 91.73±0.04 \\
ReFixMatch (2023)        & 95.06±0.01 & 95.17±0.05 & 53.88±1.07 & 72.72±0.22 & 97.85±1.23 & \textbf{98.11±0.03} & 71.40±4.21 & 94.26±0.30 \\
SoftMatch (2023)         & 94.89±0.14 & 95.04±0.09 & \underline{62.40±0.24} & 73.61±0.38 & 97.54±0.24 & 97.99±0.01 & 77.78±3.82 & 94.21±0.15 \\
EPASS (2023)             & 94.69±0.10 & 94.92±0.05 & 61.12±0.24 & \underline{74.32±0.33} & 97.02±0.02 & 97.96±0.02 & 84.39±2.48 & 94.06±1.42 \\
FreeMatch (baseline,2023)& 95.10±0.04 & 95.12±0.18 & 62.02±0.42 & 73.53±0.20 & \underline{98.03±0.02} & 98.03±0.01 & \underline{84.46±0.55} & \underline{94.37±0.15} \\
CGMatch (2025)           & \underline{95.13±0.18} & \underline{95.20±0.21} & 52.45±2.48 & 67.49±0.51 & 97.61±0.06 & 97.87±0.06 & 77.81±1.81 & -- \\
\midrule
MTSSL+FixMatch \hfill (Ours)               &93.77±0.23 &95.15±0.04 &56.05±1.71  &72.29±0.32  &98.02±0.87 &98.03±0.36   &66.47±2.29    &93.98± 0.31  \\
    Outperforms FixMatch \hfill               ~~~~~&~~~\textbf{+1.49} &~~~~+0.01 &~~~+\textbf{2.07}  &~~~~+0.32  &~~~~\textbf{+1.83} &~~~~~~+0.03    &~~~\textbf{+2.44}   &~~ +0.23 \\
    \midrule
    MTSSL+FreeMatch \hfill (Ours)               & \textbf{95.17±0.08}   &\textbf{95.21±0.10}   & \textbf{63.14±0.47}  &\textbf{74.58±0.11}    &\textbf{98.03±0.19} &\underline{98.06±0.05} &\textbf{85.41±0.82} &\textbf{94.47±0.34}\\
    Outperforms FreeMatch \hfill               & ~~~+0.07   &~~~~+0.09   & ~~~\textbf{+1.12}  &~~~~\textbf{+1.03}    &~~~~~+0 &~~~~+0.03 &~~~~\textbf{+0.95} &~~~+0.10\\
    \hline
\bottomrule
\end{tabular}
\end{adjustbox}

\end{table*}

\section{Experiment}

\subsection{Experimental Setup}
\textbf{Benchmark Datasets.} We present comprehensive experiments of MTSSL across
extensive datasets, including SVHN, CIFAR-10, CIFAR-100, and STL-10~\cite{cifar, svhn, stl10}. 
To simplify the notation, we sometimes denote the CIFAR-10 dataset with 40 labeled data as CIFAR-10 40, and similarly, denote the CIFAR-10 dataset with 250 labeled data as CIFAR-10 250. We also use this notation on CIFAR-100, SVHN, and STL-10 datasets.

\textbf{Backbone Models.} With the classic setting\cite{sohn2020fixmatch, wang2022freematch,zhao2022lassl}, for SVHN and CIFAR-10 datasets, we use WideResNet-28-2 as the encoder to generate representations. For the CIFAR-100 dataset, the encoder is WideResNet-28-8. For the STL-10 dataset, we set the encoder as WideResNet-37-2. 
The predictor is a one-layer linear network. We use the SGD optimizer to update the model's weights with a momentum of $0.9$ and a weight decay of $5 \times 10^ {-4}$. The learning rate follows a cosine decay schedule. 
For threshold $\tau$, we use Adam as the optimizer with learning rate $\eta_2=0.001$. 
We deploy our algorithm on NVIDIA GeForce RTX 3090. 

\textbf{Hyperparameters.} On all the datasets, we set $\lambda=0.02$, $g(h(\tau)) = \frac{1}{\sqrt{1-h(\tau)}}$. 
We set the smooth function as $h(\tau)=\frac{1}{1+e^{-\beta\cdot(p'_\theta(x)-\tau)}}, ~\beta=100$. 
We set the initial value of the threshold $\tau_0$ as 0.6 and update $\tau$ once every 20 iterations, that is $K=20$.

\subsection{Quantitative Results}
We show that MTSSL produces the best performance and overtakes the best SSL thresholding policy, FreeMatch  \cite{wang2022freematch}. 
Following standard SSL protocols  \cite{wang2022freematch, sosea2023marginmatch, han2025regmixmatch}, we randomly sample 40 and 250 labeled data for SVHN and CIFAR-10, and randomly sample 400 and 2500 labeled data for CIFAR-100, and randomly sample 40 and 1000 labeled data for STL-10.
Table~\ref{tabel1} reports the test accuracy of MTSSL and recent state-of-the-art SSL methods. 
Table~\ref{tabel1} demonstrates that MTSSL achieves the best performance on 7 of the total 8 settings, and still produces competitive results on the last second-best setting. 
On CIFAR-100 400, MTSSL enhances the accuracy of the baseline model FreeMatch  \cite{wang2022freematch} by \textbf{1.12\%}, achieving an accuracy of \textbf{63.14}\%. 
On CIFAR-100 2500, MTSSL enhances the FreeMatch by \textbf{1.05\%}, achieving an accuracy of \textbf{74.58}\%. 
We still want to prove MTSSL policy can reach the general good performance. 
On the baseline model FixMatch with the fixed threshold $\tau = 0.95$, with the same parameter settings, MTSSL outperforms baseline FixMatch with $\textbf{1.83}\%$ on SVHN 40. On STL-10 40 labeled data, MTSSL brings $\textbf{2.44}\%$ improvement. 
The experiment shows that the MTSSL policy can not only reach a better performance than the most competitive adaptive thresholding policy, but also can significantly improve the classical SSL algorithms with a fixed threshold. 
The key advantage of being plug‑and‑play and deployable to all SSL algorithms with the unsupervised $\widehat L_u(\tau,\theta)$ brings potential use to MTSSL for broader applications. 
In Tab.(\ref{tabelc1}), we show that MTSSL still reaches good performance with another regularization term $g(h(\tau)) = \|h(\tau)\|_2^2$. 
With the $L_2$ regularizer, MTSSL still outperforms the recent algorithms reported in Tab.~\ref{tabel1} on $7$ out of $8$ settings. 
In summary, MTSSL exhibits strong generalization as it consistently delivers promising performance. 
%Besides, key advantage of being plug‑and‑play and deployable to all common SSL algorithms brings broad potential use to MTSSL. 

\begin{figure*}[!t]
  \centering
  \includegraphics[width=1\linewidth]{ 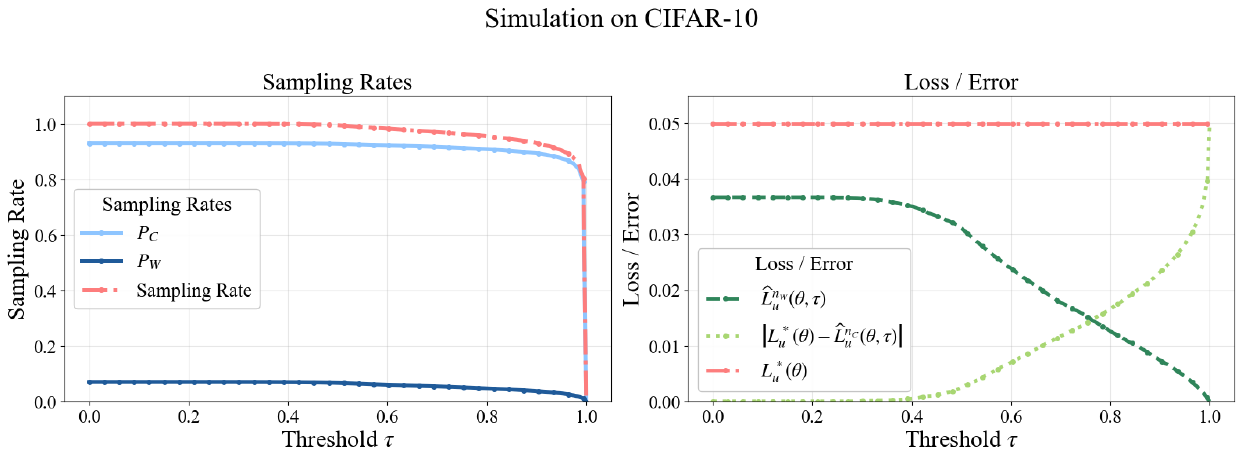}
  \vspace{0.3cm}  % 垂直间距，可调
\includegraphics[width=1\linewidth]{ 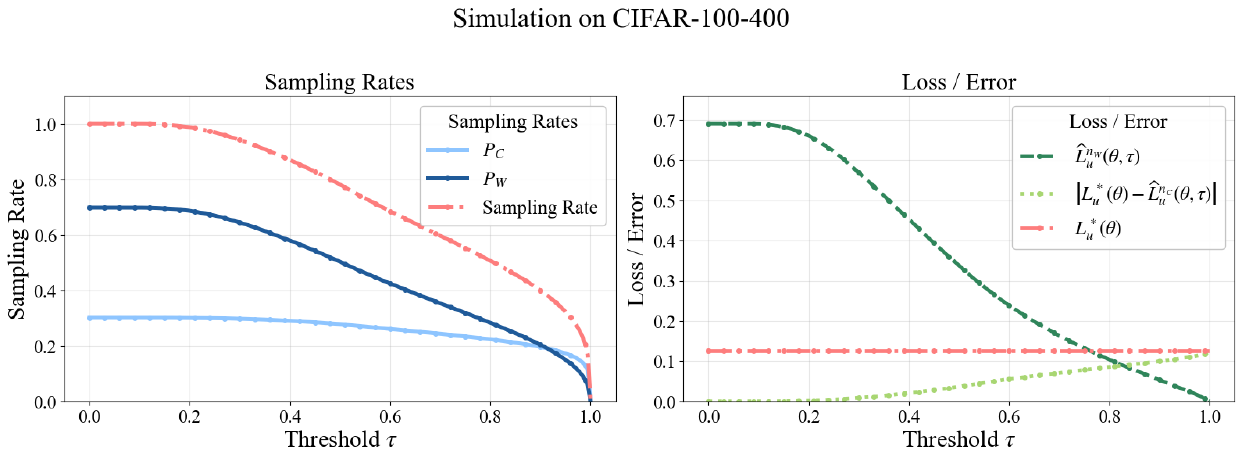}
  \caption{Simulation on the CIFAR-10 and CIFAR-100 dataset; the observation consistent with the theoretical results.}
  \label{fig3}
\end{figure*}

\subsection{Verification of Theoretical Results}
We deploy simulations on both the CIFAR-10 and CIFAR-100 datasets to validate our theoretical results and provide illustrative visualizations. 
We select the fully trained WideResNet-28-2 model shown in Tab.\ref{tabel1} from the CIFAR-10 40 setting, and similarly select the fully trained WideResNet-28-8 model for the CIFAR-100 400 setting. 
We validate the theoretical result by printing $P_C,P_W$, $|L_u^{*}(\theta) - \widehat L_u^{n_C}(\theta,\tau)|$ and $L_u^{n_W}(\theta,\tau)$ with different $\tau$ in the Fig.\ref{fig3}. 
We use $5\times10^4$ labeled data from the CIFAR-10 and CIFAR-100 datasets, and their prior labels enable us to compute the prior true loss $L^*_u(\theta)$ in Eq.(\ref{2}) approximately with sufficient prior ground-truth labels. 
%With different values of $\tau$, we plot in Fig.\ref{fig3} the value of $n_C, n_W$, the value of $|L_u^{*}(\theta) - \widehat L_u^{n_C}(\theta,\tau)|$ and $L_u^{n_W}(\theta,\tau)$. 
For \textbf{Proposition 1}, the left part of Fig.(\ref{fig3}) shows that $P_C, P_W$ monotonically decrease with the value of $\tau$ as expected. 
For \textbf{Theorem 2}, a direct analyze to Eq.(\ref{Eq5.9}) yields that for any $t,\gamma\in \mathbb{R}^+$, we have
\begin{equation*}
    \lim_{n\rightarrow +\infty}\exp\Bigl(-\frac{n^2 t^2}{2\gamma n_C}\Bigr) \rightarrow 0.
\end{equation*}
Then, with the sufficient sample size $n=5\times10^4$, we have 
\begin{equation*}
    \lim_{n\rightarrow +\infty}L_u^{n_C}(\theta,\tau) \rightarrow P_C\cdot L^*_u(\theta), ~~~~where~P_C \in [0,1].
\end{equation*}
With the fact that $P_C$ monotonically decreases with the value of $\tau$, we expect to observe that
\begin{equation*}
    \lim_{n\rightarrow +\infty}|\widehat L_u^{n_C}(\theta,\tau)- L_u^*(\theta)| = \lim_{n\rightarrow +\infty}|(1-P_C)\cdot L_u^*(\theta)|
\end{equation*}
monotonically increase with $\tau$, and 
\begin{equation*}
    \lim_{n\rightarrow +\infty,\tau\rightarrow1}|\widehat L_u^{n_C}(\theta,\tau)- L_u^*(\theta)|\rightarrow L_u^*(\theta)
\end{equation*}
as shown in the right part of Fig.\ref{fig3}. 
For \textbf{Theorem 3}, the observation in Fig.\ref{fig3} shows that the value of $\widehat L_u^{n_W}(\theta,\tau)$ monotonically decrease with $\tau$, which satisfies the conclusion that $\widehat L_u^{n_W}(\theta,\tau) = O(\frac{n_W}{n})$. 
With this inherent trade-off, the total unsupervised loss $\widehat L_u(\theta, \tau)$ in Eq.(\ref{Lu}) may take the same value with different $\tau$.
%and the same estimation error in Eq.(\ref{3}). 
This experiment directly supports the theoretical framework of this paper and reveals an interesting phenomenon in SSL: different values of $\tau$ yielding the same effect during training is possible.

\begin{figure*}[!t]
  \centering
  \begin{minipage}{0.328\linewidth}
    \centering
    \includegraphics[width=\linewidth]{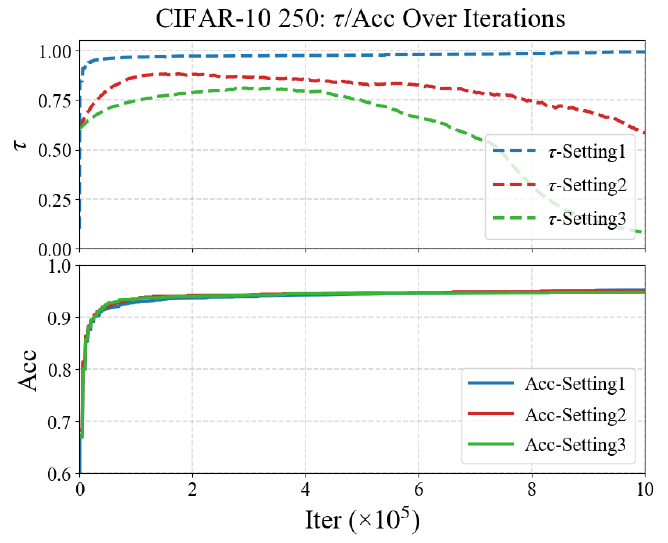}
    \par (a) $\tau$ and accuracy on CIFAR-10 with 250 labels.
  \end{minipage}
  \hfill
  \begin{minipage}{0.328\linewidth}
    \centering
    \includegraphics[width=\linewidth]{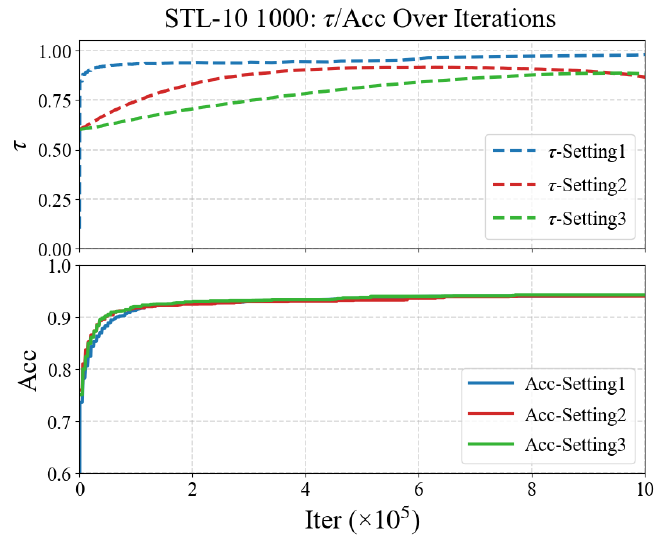}
    \par (b) $\tau$ and accuracy on STL-10 with 1000 labels.
  \end{minipage}
  \hfill
  \begin{minipage}{0.328\linewidth}
    \centering
    \includegraphics[width=\linewidth]{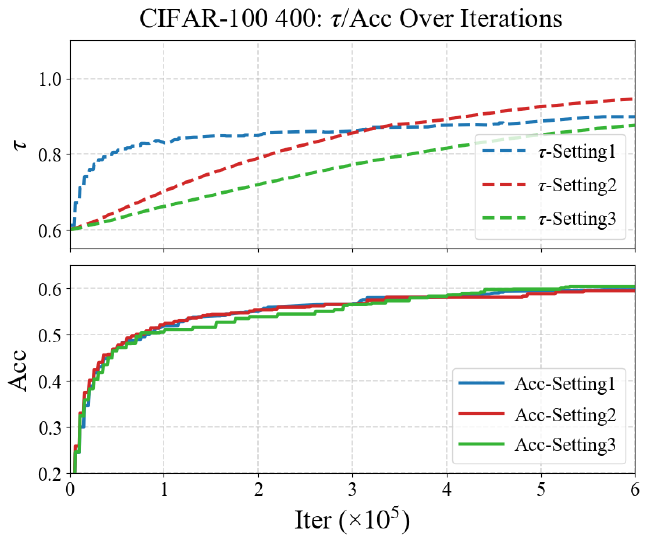}
    \par (c) $\tau$ and accuracy on CIFAR-100 with 400 labels.
  \end{minipage}
  \caption{(a,b): The accuracy almost overlaps while the values of $\tau$ are quite different. (c): With different $\tau$ values during training, accuracy can still be competitive.\label{fig:tau_acc}}
\end{figure*}

\begin{figure*}[!t]
  \centering
  \begin{minipage}{0.328\linewidth}
    \centering
    \includegraphics[width=\linewidth]{ 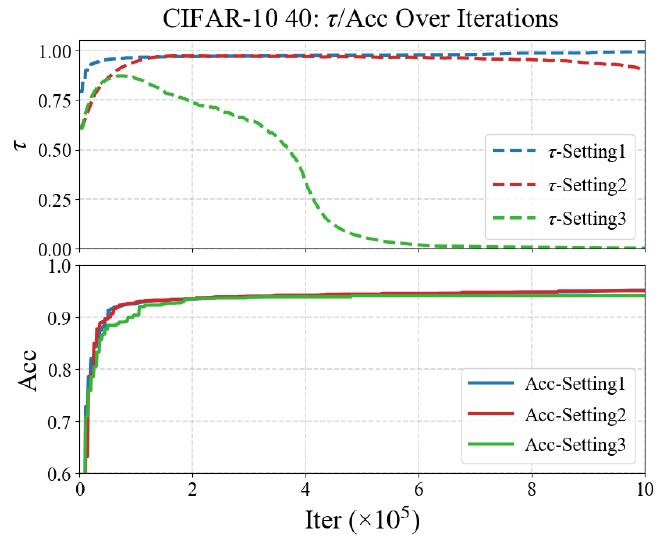}
    \par (a) Experiments on CIFAR-10 with 40 labels.
  \end{minipage}
  \hfill
  \begin{minipage}{0.328\linewidth}
    \centering
    \includegraphics[width=\linewidth]{ 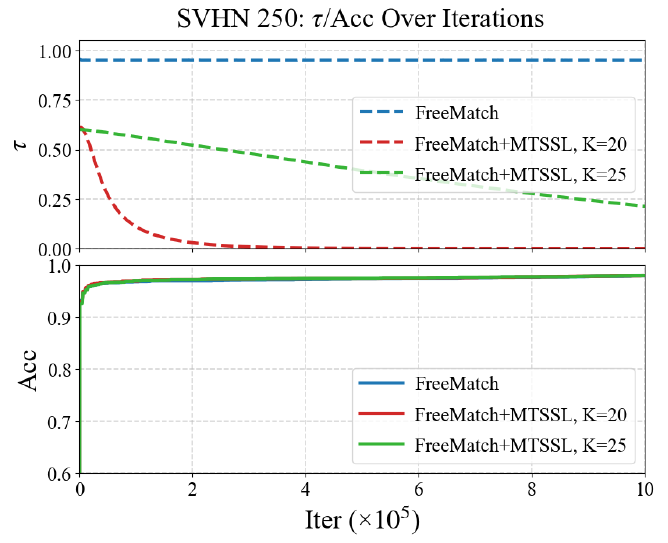}
    \par (b) Experiments on SVHN with 250 labels.
  \end{minipage}
  \hfill
  \begin{minipage}{0.328\linewidth}
    \centering
    \includegraphics[width=\linewidth]{ 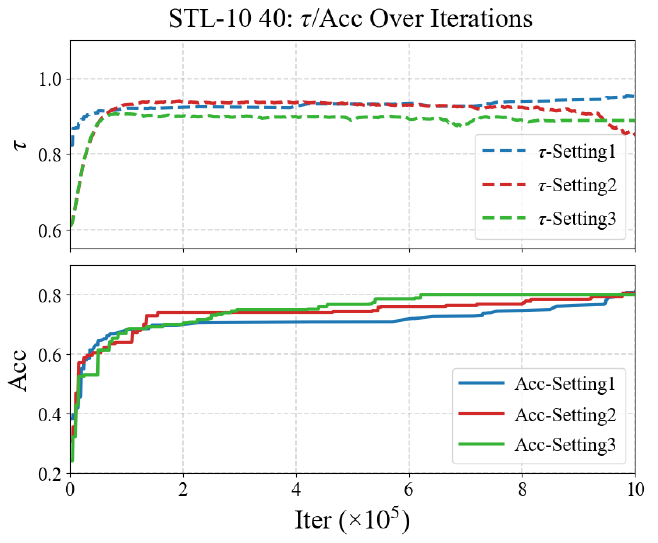}
    \par (c) Experiments on STL-10 with 40 labels.
  \end{minipage}
  \caption{The accuracy of three different experiments almost overlaps while the values of $\tau$ are quite different.  \label{appendix:extension}}
\end{figure*}

\begin{figure*}[!t]
    \centering
    \begin{minipage}{0.328\linewidth}
        \centering
        \includegraphics[width=\linewidth]{ 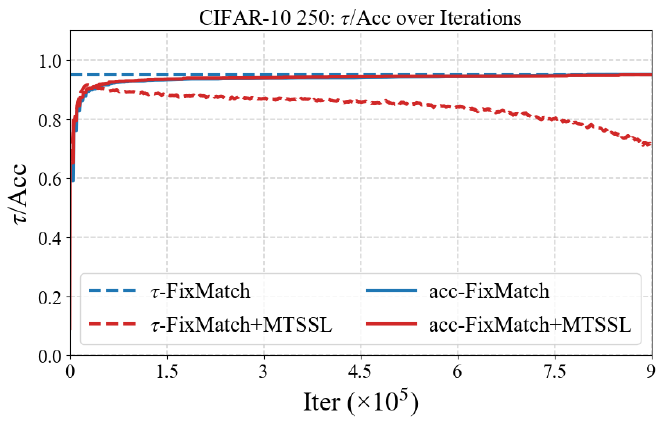}
        \par (a) Experiments on CIFAR-10 250.
        \label{fig:img1}
    \end{minipage}
    \hfill
    \begin{minipage}{0.328\linewidth}
        \centering
        \includegraphics[width=\linewidth]{ 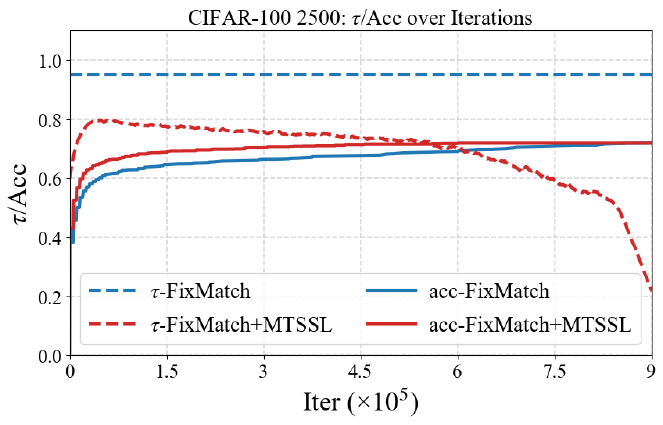}
        \par (b) Experiments on CIFAR-100 2500.
        \label{fig:img2}
    \end{minipage}
    \hfill
    \begin{minipage}{0.328\linewidth}
        \centering
        \includegraphics[width=\linewidth]{ 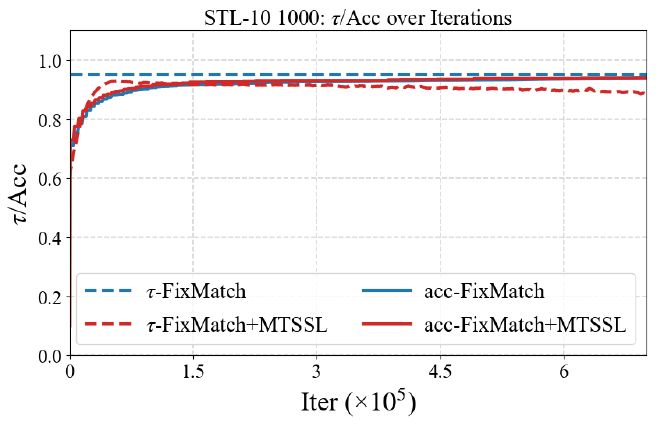}
        \par (c) Experiments on SVHN 250.
        \label{fig:img3}
    \end{minipage}
    \caption{(a), (b), (c): The accuracy of MTSSL tends to have a better increasing in the early training stages with a lower threshold $\tau$.}
    \label{fig:six_images}
\end{figure*}

\begin{figure*}[!t]
  \centering
  \begin{minipage}{0.49\linewidth}
    \centering
    \includegraphics[width=\linewidth]{ 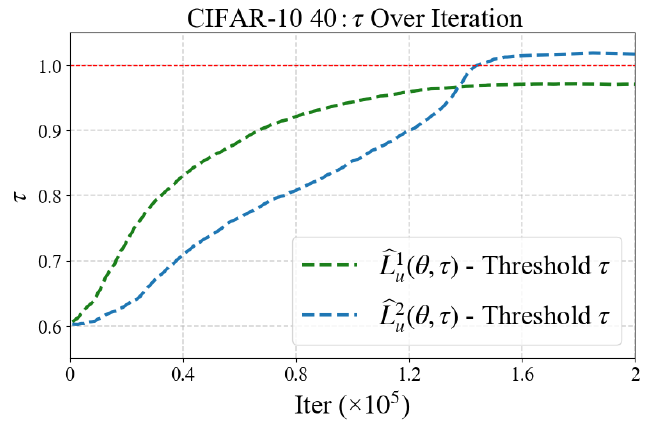}
    \par (a) Without the smoothing mapping $h(\cdot)$, $\tau$ may exceeds 1 during training.
  \end{minipage}
  \hfill
  \begin{minipage}{0.49\linewidth}
    \centering
    \includegraphics[width=\linewidth]{ 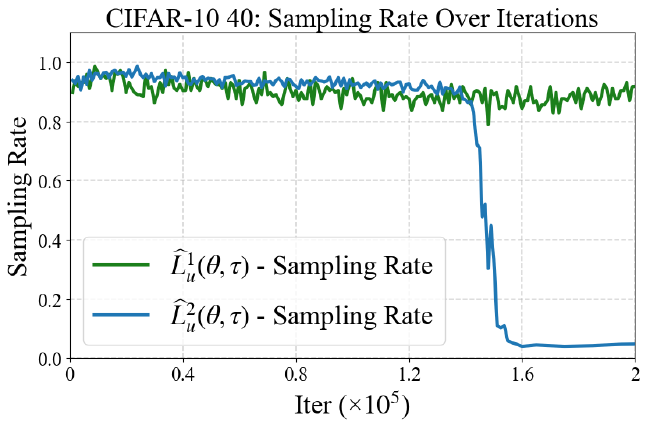}
    \par (b) The sampling rate exceeds 0 when the collapse solution happens.
  \end{minipage}
  \caption{(a) Without the smoothing mapping $h(\cdot)$, $\tau$ exceeds 1. (b) The sampling rate exceeds 0 when the collapse solution happens.}
  \label{collapse}
\end{figure*}

\begin{figure*}[!t]
  \centering
  \begin{minipage}{0.328\linewidth}
    \centering
    \includegraphics[width=\linewidth]{ 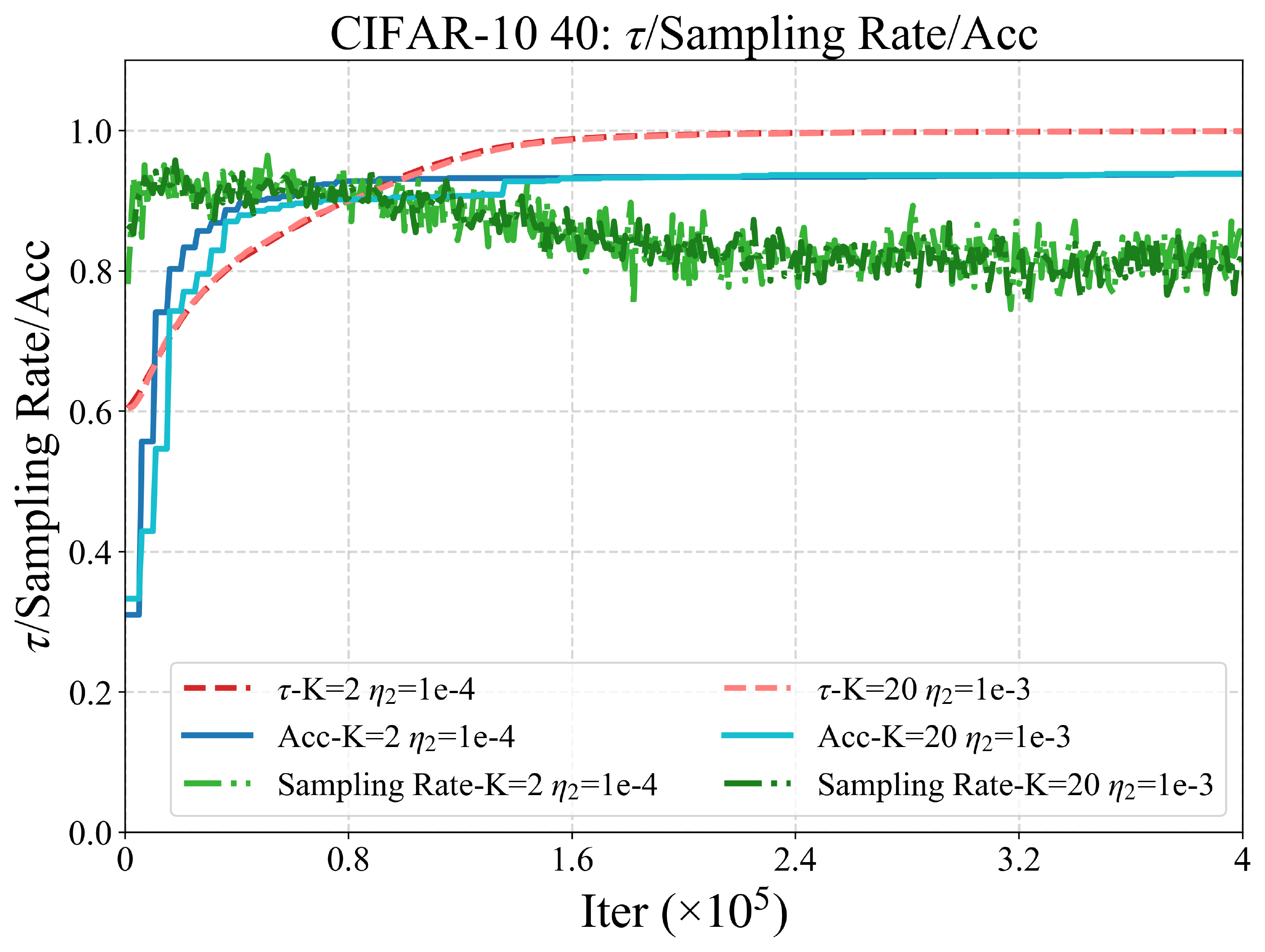}
    \par (a) Accuracy on CIFAR-10 with 40 labels.
  \end{minipage}
  \hfill
  \begin{minipage}{0.328\linewidth}
    \centering
    \includegraphics[width=\linewidth]{ 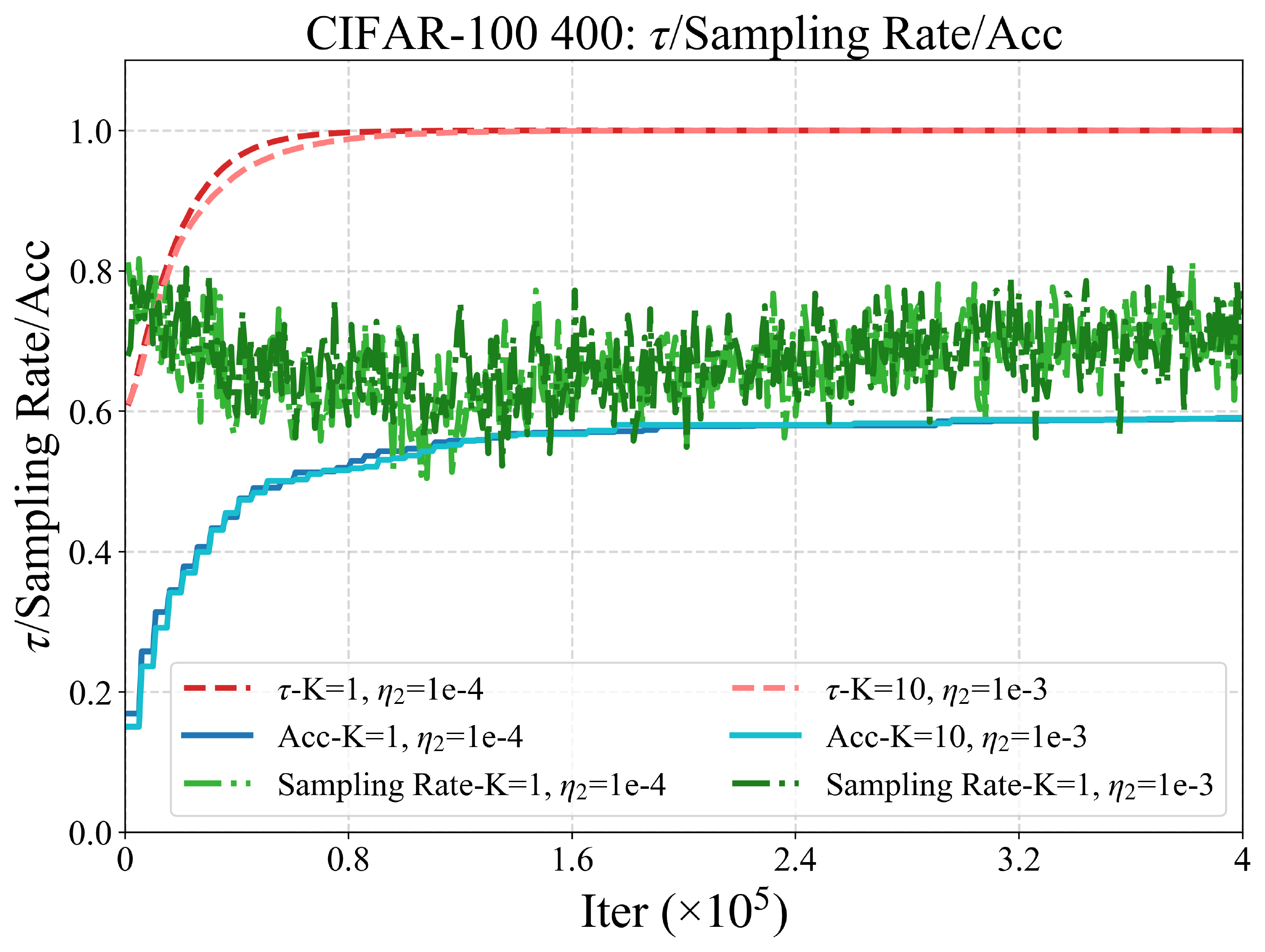}
    \par (b) Sampling rate on STL-10 with 1000 labels.
  \end{minipage}
  \hfill
  \begin{minipage}{0.328\linewidth}
    \centering
    \includegraphics[width=\linewidth]{ 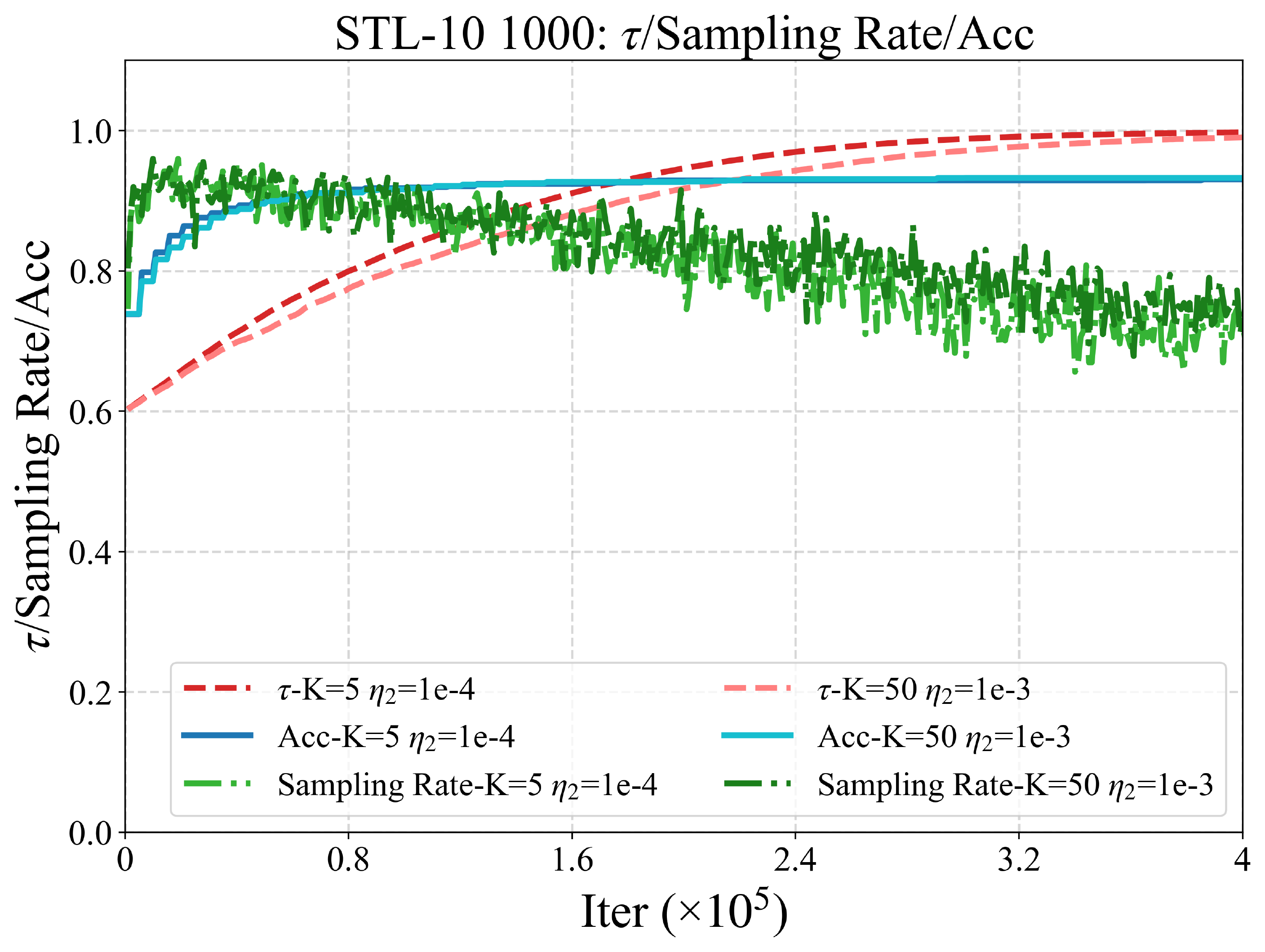}
    \par (c) Sampling rate on STL-10 with 1000 labels.
  \end{minipage}
  \caption{We replace the iteration $k$ and the learning rate $\eta_2$ for updating $\tau$, and it shows that they can reach a similar result.  \label{appendix:fig6}}

\end{figure*}

\subsection{SSL is Robust to $\tau$}
We have mentioned that different values of $\tau$ may result in the same $\widehat L_u(\theta,\tau)$ and the same estimation error $|L_u^*(\theta) - \widehat L_u(\theta,\tau)|$. 
An apparent phenomenon that we may observe is that SSL algorithms, although they have different values of $\tau$ during training, finally achieve similar accuracy. 
With FreeMatch as the baseline, we apply three parameter settings of MTSSL on each of the three datasets, yielding distinct thresholds $\tau$.
We observe in Fig.\ref{fig:tau_acc} that on CIFAR-10 250 and STL-10 1000, the accuracy curves nearly overlap even though the values of $\tau$ differ substantially. 
From Fig.\ref{fig:tau_acc}, on CIFAR-100 400, with different values of $\tau$, the accuracy curves have a similar tendency and still achieve comparable performance with different values of $\tau$ during training. 
In Fig.\ref{fig:six_images}, we compare MTSSL and the baseline model FixMatch with the fixed threshold $\tau=0.95$ and observe the same phenomenon. 
In a word, this phenomenon is widely observed across both parameter variations and model architectures, revealing a counter-intuitive finding: SSL is far less sensitive to threshold $\tau$ than we expected, which is consistent with \textbf{Theorem 2,3} and explains the counter-intuitive phenomenon in Fig.\ref{fig:introduction}. 
The observation also suggests that providing a precise value of $\tau$ case-by-case is unnecessary; even a plain gradient descent—despite its potential sub-optimality—can effectively adjust this parameter, rendering case-by-case designs \cite{wang2022freematch, zhang2021flexmatch, xu2022dash, berthelot2022adamatch} less critical.
%Although various case-by-case adaptive strategies have been designed providing precise value of $\tau$ by different willingness\cite{wang2022freematch, zhang2021flexmatch, xu2022dash, berthelot2022adamatch}, an elegant gradient descent can still tune the value of parameter $\tau$ effectively. This explains the counter-intuitive phenomenon shown in Fig.(\ref{fig:introduction}), suggesting that the design of tuning policies of $\tau$ can be relaxed in the future, for SSL may not need the precisely best value of $\tau$. 
%Besides, key advantage of being plug‑and‑play and deployable to all common SSL algorithms brings broad potential use to MTSSL.  
The parameter settings and detailed discussion can be seen in the Appendix.

In this section, we further validate this conclusion in more settings on benchmark datasets. 
As shown in Fig.\ref{appendix:extension}, this phenomenon is widespread across various datasets and settings. 
On CIFAR 10 40 and SVHN 250, the accuracy curves almost completely overlap, even when the $\tau$ values follow entirely different trajectories. 
On STL-10 40, models with different $\tau$ values finally reach the same accuracy.

\subsection{To Prevent the Collapse Solution}
In this section, we show the existence of the collapse solution, and our newly designed $\widehat L_u'(\theta,\tau)$ in Eq.(\ref{regularizer}) helps to avoid it. 
To avoid confusion, we introduce $\widehat L_u''(\theta,\tau)$ to represent the loss function in Eq.(\ref{sigmoid}), which simply replaces the indicator function in the common unsupervised loss $\widehat L_u(\theta, \tau)$ with the sigmoid function. 
That is: 
\begin{equation*}
    \widehat L_u''(\theta,\tau)=\sum_{i=1}^n  \frac{1}{1+e^{-\beta\cdot(p'_\theta(x_i)-\tau)}}\cdot H\big[p_\theta(a(x_i)),p_\theta(A(x_i)\big],
\end{equation*}
\begin{equation*}
    \widehat L_u'(\theta,\tau) = \sum_{i=1}^n  \frac{H\big[p_\theta(A(x_i)),p_\theta(a(x_i))\big]}{1+e^{-\beta\cdot(p'_\theta(x_i)-h(\tau))}}\cdot  + \lambda\cdot g(h(\tau)).
\end{equation*}
In Fig.\ref{collapse}, we show the value of $\tau$ and the sampling rate of $\widehat L_u''(\theta,\tau)$ and $\widehat L_u'(\theta,\tau)$ during training. 
We train the WRN-28-2 model on CIFAR-10 40 with the same hyper-parameters of that in Tab.(\ref{tabel1}). 
It can be seen that the $\tau$ value for $\widehat L_u'(\tau)$ can be controlled steadily lower than 1, while the $\tau$ value for $\widehat L_u''(\tau)$ can exceed 1. 
Naturally, $\widehat L_u'(\tau)$ can always provide the model with sufficient pseudo-labels, while $\widehat L_u''(\tau)$ provides very few pseudo-labels when $\tau>1$.  
It proves the existence of the collapse solution in SSL and the effectiveness of MTSSL.

\subsection{The Interval To Update $\tau$}
We have observed that robustness of $\tau$ for SSL, which may bring us some benefits in simplifying MTSSL's fine-tuning. 
That is, adjusting the parameter of MTSSL, when not severely affect the value of $\tau$, will not affect the performance very much. 
We validate this insight with the interval or stepsize of updating $\tau$. 
Intuitively, adjusting $\eta_2$, the learning rate of $\tau$, can produce similar effects to adjusting the update interval $K$, for both of them affect the update speed of the threshold $\tau$. 
%In Fig.\ref{fig:lr-tau}, we design two experiments on the CIFAR-100 400: we set $g(h(\tau))=\|h(\tau)\|_2^2$, $\lambda =0.1$. We compare the first experiment with $k=1,\eta_2=0.001$ and the second experiment with $k=10,\eta_2=0.0001$; they yield nearly overlapping $\tau$ trajectories, sampling rate, and accuracy. 
%In Fig.(\ref{appendix: fig6}), we show the experiments on CIFAR-10 40 and STL-10 1000: the same with that in Tab.(\ref{tabelc1}), we take $g(h(\tau))=||h(\tau)||^2_2$ and the weight $\lambda=0.1$.
In Fig.\ref{appendix:fig6}, for the three experiments on CIFAR-10 40, CIFAR-100 400, and STL-10 1000, we fine-tune the value of $K,\eta_2$ while keep the value of $\frac{K}{\eta_2}$ the same. 
As an example, on CIFAR-10 40, we observe that when $K=2, \eta_2=1\times 10^{-4}$ and $K=20, \eta_2=1\times 10^{-3}$, their accuracy and sampling rate almost overlap with each other, and we observe a similar phenomenon on the other 2 settings. 
This equality of $K$ and $\tau$ can be explained by \textbf{Theorem 2,3}, as the curves of $\tau$ share similar trends and differ only slightly in some intervals, so the minor difference of $\tau$ will not affect the value of $\widehat L_u(\theta,\tau)$, and thus the estimation error.
It reveals that the iteration $K$ can be replaced by a proper $\eta_2$, which can significantly simplify the hyperparameter tuning of SSL algorithms in future studies.

\begin{figure*}[!t]
  \centering
  \includegraphics[width=1\linewidth]{ 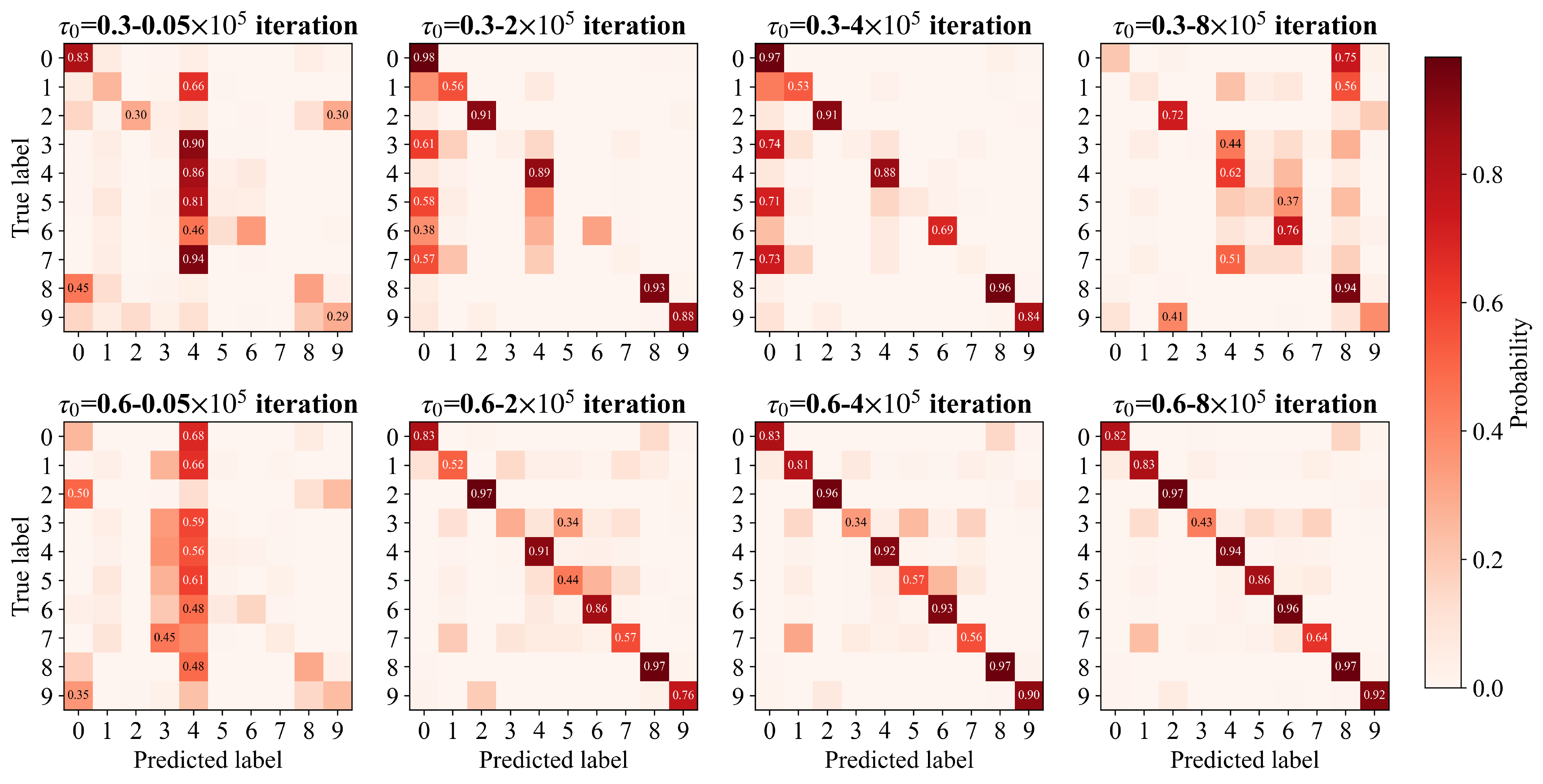}
  \caption{Confusion matrix within a batch when $\tau_0=0.3,0.6$, on $0.05,2,4,8\times10^5$ iteration. The lower $\tau_0$ tends to bring more wrong pseudo-labels in the early stages, and it may harm the prediction in the latter iterations. \label{fig:matrix}}
\end{figure*}

\subsection{Ablation Study}

\subsubsection{The Initial $\tau_0$} 
Many SSL studies point out that there is unavoidable confirmation bias in the early stages of SSL training, and that a higher threshold $\tau$ helps avoid it~\cite{arazo2020ijcnn, sohn2020fixmatch}. 
As MTSSL is a self-motivating and data-driven thresholding policy, confirmation bias in the early stage may have a greater impact on MTSSL. 
Following a common optional choice in SSL's thresholding policy~\cite{scmt, xu2022dash}, we choose the initialized $\tau_0 = 0.6$ to help MTSSL select accurate pseudo-labels while avoiding excessive rejection of low-confidence pseudo-labels~\cite{arazo2020ijcnn}. 
We compare the confusion matrices of MTSSL in Fig.\ref{fig:matrix}. 
We choose four confusion matrices in Fig.\ref{fig:matrix} from $0.05\times10^5$ to $4\times 10^5$ iterations on two experiments: we keep the other parameters the same while the initial threshold is chosen as $\tau_0 =0.3,0.6$ separately. 
%we update the $\tau$ every 20 iterations and set $\lambda = 0.1$, $g(h(\tau)) = ||h(\tau)||_2^2$, the initial $\tau_0 =0.3,0.6$ separately. 
It can be seen in the $0.05\times10^5$ and $2\times10^5$ iterations that the confusion matrix with $\tau_0=0.3$ tends to misclassify the predictions, and this tendency worsens in the $4\times10^5$ and $8\times10^5$ iterations. 
With $\tau_0=0.6$, MTSSL exhibits a better confusion matrix in the early stage, which means the model benefits from higher initial $\tau_0$.

\subsubsection{The Weight of Regularizer}

We add a regulator to control the value of $\tau$, and in this section, we show the effect of the regularizer's weight $\lambda$. 
We select the encoder WideResNet-28-2 for the CIFAR-10 250 dataset and the encoder WideResNet-37-2 for the STL-10 1000 dataset, using the FreeMatch baseline model. 
In Tab.(\ref{tabel4}), we change the value of $\lambda$ and observe the testing accuracy. 
%We keep the other parameters the same with that of in Tab.(\ref{tabel1}) and show the testing accuracy with different values of $\lambda$. 
It can be seen that with $\lambda = 0.02$, the MTSSL with regularizer $g(h(\tau)) = \frac{1}{\sqrt{1-h(\tau)}}$ reaches the best performance. 
%We point out that the different effect caused by $\lambda$ does not contradict the phenomenon shown in Fig.(\ref{fig:six_images}).
%Although different $\tau$ may bring the same effect, we still need to choose a relatively proper $\lambda$, as it may affect the gradient of $\tau$ and thus $\tau$'s growing speed. It may affect the number of $n_C, n_W$, and significantly affect MTSSL's performance in the early stage. 
%Intuitively, if MTSSL chooses worse $\tau$ values persistently during the training, it may still lead to markedly different outcomes. With a lower $\lambda$, the value of $\tau$ may tend to be higher and may provide fewer pseudo-labels; with a higher $\lambda$, the value of $\tau$ may be too low so that the MTSSL lacks the ability to select correct labels. 
%A detailed discussion about $\lambda$ and different regularizers can be seen in the appendix. 
%We emphasize that this conclusion does not contradict experiments in Fig.(\ref{fig:six_images}): although different $\tau$ may yield the same $\widehat L_u(\tau,\theta)$, the trade-off is heavily influenced by $P_C, P_W$ and varies across different models and datasets. 
Intuitively, persistently choosing worse $\tau$ may affect the results: lower $\lambda$ yields fewer pseudo-labels, while higher $\lambda$ risks incorrect predictions. 
A detailed discussion about $\lambda$ can be seen in the appendix.

\subsubsection{The Shape of Sigmoid Function}

In MTSSL, we use a sigmoid function to smooth the unsupervised loss $\widehat L_u'(\theta,\tau)$  \cite{byod, nestrov} so that MTSSL is differentiable. 
It comes to us that the value of $\beta$ affects the shape of the sigmoid function; with a bigger $\beta$, the sigmoid function becomes closer to the indicator function. 
In Tab.\ref{tabel5}, we show the accuracy with $\beta = 1, 10, 100, 1000,10000$ on CIFAR-10 40. 
It can be seen that when we have $\theta=100$, it achieves the best result. At the same time, when $\beta$ is too large or too small(eg, $\beta$=1, 10000), MTSSL gets worse. 
%It complies with our intuition that when $\theta$ is too large, only very few input data with excessively high confidence can affect the differentiation of $\tau$. 
%When $\theta$ is small, too many inputs with comparatively small confidence may affect the value of $\tau$. 
%When $\theta=1, 10$, we observe that $\tau$ exhibits a monotonically decreasing trend during the training. We explain this by noticing that in \textbf{Proposition 4}, the differentiation of $\tau$ is directly affect with the value of $\theta$, and a lower $\theta$ may result in $\frac{\partial \widehat L_u(\theta,\tau)}{\partial \tau}<0$, thus encourages the threshold $\tau$ to decrease during the training monotonically. 
To be on the safe side, we recommend using a proper $\beta$ that approximates an indicator function suitably.

\begin{table*}[t] 
 \caption{Ablation study with the weighting parameter $\lambda$.}
  \label{tabel4}
  \centering
\fontsize{8}{11}\selectfont
  \begin{tabular}{c|c|c|c|c|c|c}
    \toprule
    \hline
     $g(h(\tau))=\frac{1}{\sqrt{1-h(\tau)}}$ & $\lambda=0.003$  & $\lambda=0.005$  & $\lambda=0.01$ & $\lambda=0.02$  &$\lambda=0.05$   &$\lambda=0.1$  \\
    \midrule
    CIFAR-10 250   &94.89   &95.03    &95.09 &\textbf{95.23}   &94.85    &94.76      \\
    \midrule
    STL-10 1000   &93.78  &93.90    &94.10  &\textbf{94.44}  &94.31    &94.25      \\
    \hline
    \bottomrule 
  \end{tabular}
\end{table*}

\begin{table*}[t] 
 \caption{Ablation study with the value of $\beta$.}
  \label{tabel5}
  \centering
\fontsize{9.1}{11.5}\selectfont
  \begin{tabular}{c|c|c|c|c|c}
    \toprule
    \hline
     Value of $\beta$& $\beta=1$ & $\beta=10$  & $\beta=100$  & $\beta=1000$ & $\beta=10000$   \\
    \midrule
    CIFAR-10 40   &94.21  &94.74   &\textbf{95.16}    &95.02 &94.87      \\
    \hline
    \bottomrule 
  \end{tabular}
\end{table*}

\subsubsection{ $L_2$ Regularizer: Quantitative Results}

\begin{table*}
\caption{Performance comparison ($g(h(\tau)) = \|\tau\|^2_2$) on CIFAR10, CIFAR100, SVHN, and STL10. 
We report mean accuracy (\%) and standard deviation over 5 runs. 
\textbf{Bold} and \underline{underlined} numbers indicate the best and second-best results, respectively.}
\label{tabelc1}
\centering
\footnotesize   % 正常字号，缩放后会等比例缩小，但不会像4.5pt那么夸张
\setlength{\tabcolsep}{5pt}  % 列间距稍大，避免缩放后数字贴在一起
\renewcommand{\arraystretch}{1.5}
\begin{adjustbox}{width=\textwidth}   % 整体缩放到页面宽度
\begin{tabular}{@{}l cccccccc@{}}
\toprule
\hline
\multirow{2}{*}{Method} & \multicolumn{2}{c}{CIFAR-10} & \multicolumn{2}{c}{CIFAR-100} & \multicolumn{2}{c}{SVHN} & \multicolumn{2}{c}{STL-10} \\
\cmidrule(r){2-3} \cmidrule(r){4-5} \cmidrule(r){6-7} \cmidrule(r){8-9}
& 40 labels & 250 labels & 400 labels & 2500 labels & 40 labels & 250 labels & 40 labels & 1000 labels \\
\midrule
FixMatch (2020)          & 92.53±0.28 & 95.14±0.05 & 53.58±0.82 & 71.97±0.16 & 96.19±1.18 & 98.03±0.01 & 64.03±4.14 & 93.75±0.33 \\
FreeMatch (baseline,2023)& 95.10±0.04 & 95.12±0.18 & 62.02±0.42 & 73.53±0.20 & \underline{98.03±0.02} & 98.03±0.01 & \underline{84.46±0.55} & \underline{94.37±0.15} \\
\midrule
MTSSL+Fixmatch (Ours)   & 93.77±0.23 & 95.15±0.04 & 55.09±0.75 & 71.99±0.15 & 98.02±0.87 & 98.03±0.97 & 66.67±3.47 & 93.98±0.31 \\
Outperforms Fixmatch    & \textbf{+1.49} & +0.01 & \textbf{+1.11} & +0.02 & \textbf{+1.83} & +0 & \textbf{+2.64} & +0.23 \\
\midrule
MTSSL+Freematch (Ours)  & 95.15±0.08 & 95.21±0.10 & \textbf{62.74±0.94} & \textbf{74.46±0.19} & 98.03±0.19 & 98.06±0.05 & \textbf{84.91±0.61} & 94.48±0.24 \\
Outperforms Freematch   & +0.05 & +0.09 & \textbf{+0.72} & \textbf{+0.93} & +0 & +0.03 & \textbf{+0.45} & +0.11 \\
\hline
\bottomrule
\end{tabular}
\end{adjustbox}
\end{table*}

\iffalse
\begin{table*}[t] 
 \caption{Ablation study with the weighting parameter $\lambda$.}
  \label{tabelc2}
  \centering
\fontsize{9.1}{12}\selectfont
  \begin{tabular}{c|c|c|c|c|c|c}
    \toprule
    \hline
     $g(h(\tau))=\|h(\tau)\|_2^2$ & $\lambda=0.01$  & $\lambda=0.05$  & $\lambda=0.1$ & $\lambda=0.3$  &$\lambda=0.5$   &$\lambda=1$  \\
    \midrule
    CIFAR-10 250   &95.03   &95.11    &\textbf{95.31 } &94.89  &94.45    &94.15      \\
    \hline
    \bottomrule 
  \end{tabular}
\end{table*}
\fi

We show that MTSSL exhibits excessive performance with different regularizers.
In Tab.(\ref{tabelc1}), we set $g(h(\tau)) = \|h(\tau)\|_2^2$, $\lambda = 0.1$ while keeping other hyperparameters the same as those in Tab.(\ref{tabel1}). 
It can be seen that MTSSL still outperforms the fixed-threshold SSL algorithm, FixMatch, by \textbf{2.64\%} on STL-10 with 40 labels. On CIFAR-10 with 400 labels and CIFAR-100 with 400 labels, it surpasses FixMatch by \textbf{1.49\%} and \textbf{1.83\%}, respectively, sufficiently improving the performance of the baseline model FixMatch with a fixed threshold. 
For the best existing thresholding policy, FreeMatch, MTSSL further improves it by \textbf{$0.72\%$} and \textbf{$0.93\%$} on CIFAR-100 with 400 labels and CIFAR-100 with 2500 labels, respectively. 
With the $L_2$ regularizer, MTSSL still outperforms the recent algorithms reported in Tab.~\ref{tabelc1} on $7$ out of $8$ settings. 
In summary, even with the common $L_2$ regularizer, MTSSL can still stably outperform the best existing thresholding policy.
%We also deploy the ablation study to the weight $\lambda$ in Tab.\ref{tabelc2}. When we take $g(h(\tau))=\|h(\tau)\|_2^2$, $\lambda=0.1$ yields the best performance. 
%We emphasize that MTSSL is a flexible parameter-tuning strategy, allowing the free construction of task-specific regularization terms according to the needs.

\section{Conclusion} 

We are interested in explaining why different adaptive thresholding policies can make improvements to SSL with quite different motivations and $\tau$ values. 
To address this, we first generate a statistical framework, pointing out that the empirical unsupervised loss $\widehat L_u(\theta,\tau)$ can be divided into two parts affected by the correct pseudo-labels and the confirmation-bias, separately. 
In \textbf{Proposition 1} and \textbf{Theorem 2,3}, we further explain threshold $\tau$ affects the estimation error of $|\widehat L_u(\theta,\tau) - L_u^*(\theta)|$ by balancing the number of correct and wrong pseudo-labels. 
With this, we design MTSSL to update $\tau$ with elegant back-propagation, and expect it to still surpass the existing SSL thresholding policies, for SSL algorithms may reach the same unsupervised loss with different values of $\tau$. 
%We observe and explain the collapse solution in SSL for the first time: $\tau$ will sharply increase to $1$ and reject all the inputs, and avoid it successfully in the design of MTSSL.
Extensive experiments show the competitive performance of the MTSSL. We also observe that SSL achieves almost overlapping performance during training with quite different values of $\tau$. 
We reveal both theoretically and experimentally that the SSL algorithms are not very sensitive to the value of $\tau$, and thus we can simplify the selection of $\tau$ in future SSL algorithms.

\newpage

\iffalse
\begin{IEEEbiographynophoto}{John Doe}
Use $\backslash${\tt{begin\{IEEEbiographynophoto\}}} and the author name as the argument followed by the biography text.
\end{IEEEbiographynophoto}
\fi

\vfill


\begin{thebibliography}{1}
\bibliographystyle{IEEEtran}



\bibitem{iscen2019label}
A. Iscen, G. Tolias, Y. Avrithis, and O. Chum, ``Label propagation for deep semi-supervised learning,'' \textit{IEEE / CVF Conference on Computer Vision and Pattern Recognition (CVPR)}, pp. 5070--5079, 2019.

\bibitem{li2020density}
S. Li, B. Liu, D. Chen, Q. Chu, L. Yuan, and N. Yu, ``Density-aware graph for deep semi-supervised visual recognition,'' \textit{IEEE / CVF Conference on Computer Vision and Pattern Recognition (CVPR)}, pp. 13400--13409, 2020.

\bibitem{zhao2022lassl}
Z. Zhao, L. Zhou, L. Wang, Y. Shi, and Y. Gao, ``LASSL: Label-guided self-training for semi-supervised learning,'' \textit{AAAI Conference on Artificial Intelligence (AAAI)}, vol. 36, pp. 9208--9216, 2022.

\bibitem{chen2020big}
T. Chen, S. Kornblith, K. Swersky, M. Norouzi, and G. E. Hinton, ``Big self-supervised models are strong semi-supervised learners,'' \textit{Conference on Neural Information Processing Systems (NeurIPS)}, 2020.

\bibitem{berthelot2019mixmatch}
D. Berthelot, N. Carlini, I. Goodfellow, N. Papernot, A. Oliver, and C. A. Raffel, ``Mixmatch: A holistic approach to semi-supervised learning,'' \textit{Conference on Neural Information Processing Systems (NeurIPS)}, vol. 32, 2019.

\bibitem{berthelot2019remixmatch}
D. Berthelot, N. Carlini, E. D. Cubuk, A. Kurakin, K. Sohn, H. Zhang, and C. Raffel, ``Remixmatch: Semi-supervised learning with distribution alignment and augmentation anchoring,'' \textit{International Conference on Learning Representations (ICLR)}, 2019.

\bibitem{sohn2020fixmatch}
K. Sohn, D. Berthelot, N. Carlini, Z. Zhang, H. Zhang, C. A. Raffel, E. D. Cubuk, A. Kurakin, and C. L. Li, ``Fixmatch: Simplifying semi-supervised learning with consistency and confidence,'' \textit{Conference on Neural Information Processing Systems (NeurIPS)}, vol. 33, pp. 596--608, 2020.

\bibitem{miyato2018virtual}
T. Miyato, S. I. Maeda, M. Koyama, and S. Ishii, ``Virtual adversarial training: a regularization method for supervised and semi-supervised learning,'' \textit{IEEE Transactions on Pattern Analysis and Machine Intelligence (TPAMI)}, vol. 41, no. 8, pp. 1979--1993, 2018.

\bibitem{tarvainen2017mean}
A. Tarvainen and H. Valpola, ``Mean teachers are better role models: Weight-averaged consistency targets improve semi-supervised deep learning results,'' \textit{Conference on Neural Information Processing Systems (NeurIPS)}, vol. 30, 2017.

\bibitem{rasmus2015semi}
A. Rasmus, M. Berglund, M. Honkala, H. Valpola, and T. Raiko, ``Semi-supervised learning with ladder networks,'' \textit{Conference on Neural Information Processing Systems (NeurIPS)}, vol. 28, 2015.

\bibitem{lee2013pseudo}
D. H. Lee, ``Pseudo-label: The simple and efficient semi-supervised learning method for deep neural networks,'' \textit{International Conference on Machine Learning (ICML)}, vol. 3, no. 2, pp. 896, 2013.

\bibitem{douze2018low}
M. Douze, A. Szlam, B. Hariharan, and H. Jégou, ``Low-shot learning with large-scale diffusion,'' \textit{IEEE / CVF Conference on Computer Vision and Pattern Recognition (CVPR)}, pp. 3349--3358, 2018.

\bibitem{zheng2022simmatch}
M. Zheng, S. You, L. Huang, F. Wang, C. Qian, and C. Xu, ``Simmatch: Semi-supervised learning with similarity matching,'' \textit{IEEE / CVF Conference on Computer Vision and Pattern Recognition (CVPR)}, pp. 14471--14481, 2022.

\bibitem{wang2022freematch}
Y. Wang, H. Chen, Q. Heng, W. Hou, Y. Fan, Z. Wu, et al., ``Freematch: Self-adaptive thresholding for semi-supervised learning,'' \textit{International Conference on Learning Representations (ICLR)}, 2022.

\bibitem{guo2022adsh}
L. Z. Guo and Y. F. Li, ``Class-imbalanced semi-supervised learning with adaptive thresholding,'' \textit{International Conference on Machine Learning (ICML)}, pp. 8082--8094, 2022.

\bibitem{li2021comatch}
J. Li et al., ``Comatch: Semi-supervised learning with contrastive graph regularization,'' \textit{IEEE/CVF International Conference on Computer Vision (ICCV)}, pp. 9475--9484, 2021.

\bibitem{azizyan2013density}
M. Azizyan, A. Singh, and L. Wasserman, ``Density-sensitive semisupervised inference,'' \textit{Annals of Statistics (ANNALS)}, vol. 41, no. 2, pp. 751--771, 2013.

\bibitem{chen2023softmatch}
H. Chen, R. Tao, Y. Fan, Y. Wang, J. Wang, B. Schiele, and M. Savvides, ``Softmatch: Addressing the quantity-quality trade-off in semi-supervised learning,'' \textit{International Conference on Learning Representations (ICLR)}, vol. 8, 2023.

\bibitem{yang2023shrinking}
L. Yang, Z. Zhao, L. Qi, Y. Qiao, Y. Shi, and H. Zhao, ``Shrinking class space for enhanced certainty in semi-supervised learning,'' \textit{IEEE/CVF International Conference on Computer Vision (ICCV)}, pp. 16187--16196, 2023.

\bibitem{nguyen2023boosting}
K. B. Nguyen, ``Boosting Semi-Supervised Learning by bridging high and low-confidence prediction,'' \textit{IEEE/CVF International Conference on Computer Vision (ICCV)}, pp. 1028--1038, 2023.

\bibitem{zhou2003learning}
D. Zhou, O. Bousquet, T. Lal, J. Weston, and B. Schölkopf, ``Learning with local and global consistency,'' \textit{Conference on Neural Information Processing Systems (NeurIPS)}, vol. 16, 2003.

\bibitem{han2025regmixmatch}
H. Han, J. Yuan, and C. Wei, ``RegMixMatch: Optimizing Mixup Utilization in Semi-Supervised Learning,'' \textit{AAAI Conference on Artificial Intelligence (AAAI)}, vol. 39, no. 16, pp. 17032--17040, 2025.

\bibitem{xu2022dash}
Y. Xu, L. Shang, and J. Ye, ``Dash: Semi-supervised learning with dynamic thresholding,'' \textit{International Conference on Machine Learning (ICML)}, pp. 11525--11536, 2021.

\bibitem{li2020dividemix}
J. Li, R. Socher, and S. Hoi, ``Dividemix: Learning with noisy labels as semi-supervised learning,'' \textit{International Conference on Learning Representations (ICLR)}, 2020.

\bibitem{arazo2020ijcnn}
E. Arazo, D. Ortego, and P. Albert, ``Pseudo-labeling and confirmation bias in deep semi-supervised learning,'' \textit{Proceedings of the International Joint Conference on Neural Networks (IJCNN)}, pp. 1--8, 2020.

\bibitem{zhang2021flexmatch}
B. Zhang, Y. Wang, and W. Hou, ``Flexmatch: Boosting semi-supervised learning with curriculum pseudo labeling,'' \textit{Conference on Neural Information Processing Systems (NeurIPS)}, vol. 34, pp. 18408--18419, 2021.

\bibitem{sosea2023marginmatch}
T. Sosea and C. Caragea, ``MarginMatch: Improving semi-supervised learning with pseudo-margins,'' \textit{IEEE / CVF Conference on Computer Vision and Pattern Recognition (CVPR)}, pp. 15773--15782, 2023.

\bibitem{berthelot2022adamatch}
D. Berthelot, R. Roelofs, and K. Sohn, ``Adamatch: A unified approach to semi-supervised learning and domain adaptation,'' \textit{International Conference on Learning Representations (ICLR)}, vol. 7, 2022.

\bibitem{chen2023boosting}
Y. Chen, X. Tan, and B. Zhao, ``Boosting Semi-Supervised Learning by Exploiting All Unlabeled Data,'' \textit{IEEE / CVF Conference on Computer Vision and Pattern Recognition (CVPR)}, pp. 7548--7557, 2023.

\bibitem{li2023instant}
M. Li, R. Wu, and H. Liu, ``InstanT: Semi-supervised Learning with Instance-dependent Thresholds,'' \textit{Conference on Neural Information Processing Systems (NeurIPS)}, vol. 36, pp. 2922--2938, 2023.

\bibitem{xie2020uda}
Q. Xie, Z. Dai, E. Hovy, et al., ``Unsupervised data augmentation for consistency training,'' \textit{Neurips}, 2020.

\bibitem{yang2020rethinking}
Y. Yang and Z. Xu, ``Rethinking the Value of Labels for Improving Class-Imbalanced Learning,'' \textit{Conference on Neural Information Processing Systems (NeurIPS)}, vol. 33, pp. 19290--19301, 2020.

\bibitem{duchi2011adagard}
J. Duchi, E. Hazan, and Y. Singer, ``Adaptive subgradient methods for online learning and stochastic optimization,'' \textit{Journal of Machine Learning Research (JMLR)}, vol. 12, no. 7, 2011.

\bibitem{2022mlr}
J. Shu, Y. Zhu, and Q. Zhao, ``Mlr-snet: Transferable lr schedules for heterogeneous tasks,'' \textit{IEEE Transactions on Pattern Analysis and Machine Intelligence (TPAMI)}, vol. 45, no. 3, pp. 3505--3521, 2022.

\bibitem{xiao2021aistat}
T. Xiao, X. Y. Zhang, and H. Jia, ``Semi-supervised learning with meta-gradient,'' \textit{International Conference on Artificial Intelligence and Statistics (AISTATS)}, pp. 73--81, 2021.

\bibitem{zhang2023MLH}
L. Zhang and B. J. Mortazavi, ``Semi-supervised meta-learning for multi-source heterogeneity in time-series data,'' \textit{Machine Learning for Healthcare Conference}, pp. 923--941, 2023.

\bibitem{byod}
S. P. Boyd and L. Vandenberghe, ``Convex optimization,'' \textit{Cambridge university press}, 2004.  % 书籍，无页码

\bibitem{nestrov}
Y. Nesterov, ``Introductory lectures on convex optimization: A basic course,'' \textit{Springer Science \& Business Media}, 2013.  % 书籍

\bibitem{Goodfellow}
I. J. Goodfellow, J. Shlens, and C. Szegedy, ``Explaining and harnessing adversarial examples,'' \textit{International Conference on Learning Representations (ICLR)}, 2014.

% 以下是从 ###ICML_2026_ 之后的有效条目（跳过乱码和孤立括号）

\bibitem{gaussian_augmentation}
A. Ghosh and A. H. Thiery, ``On data-augmentation and consistency-based semi-supervised learning,'' \textit{arXiv:2101.06967}, 2021.

\bibitem{cifar}
A. Krizhevsky and G. Hinton, ``Learning multiple layers of features from tiny images,'' \textit{University of Toronto}, 2009.  % 技术报告

\bibitem{svhn}
Y. Netzer, T. Wang, A. Coates, A. Bissacco, B. Wu, and A. Y. Ng, ``Reading digits in natural images with unsupervised feature learning,'' \textit{NeurIPS Workshop on Deep Learning and Unsupervised Feature Learning}, 2011.

\bibitem{stl10}
A. Coates, A. Ng, and H. Lee, ``An analysis of single-layer networks in unsupervised feature learning,'' \textit{Proceedings of the fourteenth international conference on artificial intelligence and statistics}, pp. 215--223, 2011.

\bibitem{scmt}
F. Xiong, J. Tian, and Z. Hao, ``SCMT: Self-Correction Mean Teacher for Semi-supervised Object Detection,'' \textit{International Joint Conference on Artificial Intelligence (IJCAI)}, vol. 30, pp. 1488--1494, 2022.

\bibitem{sub-gaussian1}
K. Scaman, C. Malherbe, and L. Santos, ``Convergence Rates of Non-Convex Stochastic Gradient Descent Under a Generic Łojasiewicz Condition and Local Smoothness,'' \textit{International Conference on Machine Learning (ICML)}, 2022.

\bibitem{sub-gaussian2}
O. Yüksel and N. Flammarion, ``On the Sample Complexity of Next-Token Prediction,'' \textit{International Conference on Artificial Intelligence and Statistics (AISTATS)}, pp. 694--702, 2025.

\bibitem{sub-gaussian3}
S. Kanai, M. Yamada, and S. Yamaguchi, ``On the Sample Complexity of Next-Token Prediction,'' \textit{Proceedings of the International Joint Conference on Neural Networks (IJCNN)}, 2021.

\bibitem{ijcv1}
Y. Fan, A. Kukleva, and B. Schiele, ``Revisiting Consistency Regularization for Semi-Supervised Learning,'' \textit{International Journal of Computer Vision (IJCV)}, vol. 131, pp. 626--643, 2023.

\bibitem{ijcv2}
K. Zhou, C. Loy, and Z. Liu, ``Semi-Supervised Domain Generalization with Stochastic StyleMatch,'' \textit{International Journal of Computer Vision (IJCV)}, vol. 131, pp. 2377--2387, 2023.

\bibitem{ijcv3}
G. Zhao, G. Li, and Y. Qin, ``Exploration and Exploitation of Unlabeled Data for Open-Set Semi-supervised Learning,'' \textit{International Journal of Computer Vision (IJCV)}, vol. 132, pp. 5888--5904, 2024.

\bibitem{ijcv4}
Q. Yang, Z. Chen, and Z. Peng, ``Relation-Guided Versatile Regularization for Federated Semi-Supervised Learning,'' \textit{International Journal of Computer Vision (IJCV)}, vol. 133, pp. 3312--3326, 2025.

\end{thebibliography}
\end{document}